\documentclass[journal]{IEEEtran}
\usepackage{url}
\usepackage{graphicx} 
\usepackage{multirow}
\usepackage{amsmath}
\usepackage[ruled]{algorithm2e}   
\usepackage{algpseudocode}
\usepackage[T1]{fontenc}
\usepackage{amssymb}
\usepackage{threeparttable}
\usepackage{multirow}
\usepackage{textcomp,booktabs}
\usepackage[usenames]{color}
\usepackage{colortbl}
\usepackage{bm}
\usepackage{helvet}  
\usepackage{courier}  
\usepackage{url}  
\usepackage{graphicx}  
\usepackage{amsmath}
\usepackage{amssymb}
\usepackage{multirow}
\usepackage{lineno}
\usepackage{array}
\usepackage{lineno}
\usepackage{bm}
\usepackage{subfig}
\usepackage{enumerate}
\newcommand{\eg}{\textit{e.g.}}
\newcommand{\al}{\textit{et al. }}

\SetKwInput{KwInput}{Input}        
\SetKwInput{KwOutput}{Output}
\ifCLASSINFOpdf
\else
\fi
\begin{document}

\title{Exploiting Web Images for Fine-Grained Visual Recognition by Eliminating Noisy Samples and Utilizing Hard Ones}

\author{
Huafeng Liu,
Chuanyi Zhang,
Yazhou Yao,
Xiu-Shen Wei,       
Fumin Shen,
Zhenmin Tang,
and Jian Zhang        
\thanks{H. Liu, C. Zhang, Y. Yao, X. Wei, and Z. Tang are with the School of Computer Science and Engineering, Nanjing University of Science and Technology, Nanjing, China. H. Liu and C. Zhang are the co-first authors, corresponding author: Yazhou Yao, Email: Yazhou.Yao@njust.edu.cn.}	
\thanks{F. Shen is with the School of Computer Science and Engineering, University of Electronic Science and Technology of China, Chengdu, China.}
\thanks{J. Zhang is with the Global Big Data Technologies Center, University of Technology Sydney, Australia.}
}

\maketitle

\begin{abstract}
	
Labeling objects at a subordinate level typically requires expert knowledge, which is not always available when using random annotators. As such, learning directly from web images for fine-grained recognition has attracted broad attention. However, the presence of label noise and hard examples in web images are two obstacles for training robust fine-grained recognition models. Therefore, in this paper, we propose a novel approach for removing irrelevant samples from real-world web images during training, while employing useful hard examples to update the network. Thus, our approach can alleviate the harmful effects of irrelevant noisy web images and hard examples to achieve better performance. Extensive experiments on three commonly used fine-grained datasets demonstrate that our approach is far superior to current state-of-the-art web-supervised methods. The data and source code of this work have been made publicly available at: {\url{https://github.com/NUST-Machine-Intelligence-Laboratory/Advanced-Softly-Update-Drop}}.

\end{abstract}


\IEEEpeerreviewmaketitle

\section{Introduction}

Deep neural networks (DNNs) have achieved impressive results on many computer vision tasks (\eg, image retrieval \cite{zhang2020deep,hu2020pyretri,wang2020set,yao2019dynamically}, image classification \cite{li2020field,xie2019attentive,yao2018extracting,xie2020region,sun2020exploiting}, image segmentation \cite{zhoutarget,zhou2020motion,chen2020classification,lu2020hsi,luo2019segeqa}) due to the availability of large-scale image datasets, such as ImageNet \cite{imagenet}. However, fine-grained visual classification (FGVC) remains challenging \cite{zhang2020web}. Dividing a category into subclasses exponentially increases the required number of labels, making it a labor-intensive and time-consuming task. Moreover, fine-grained annotation usually requires domain-specific expert knowledge, which exacerbates the labeling problem. To reduce the cost of manual labeling, several works have focused on the semi-supervised paradigm \cite{xu2015augmenting,niu2018webly,cui2016fine}. However, these works still inevitably involve some form of human intervention and thus remain labor-intensive.

In contrast to manually labeled image datasets, web images are a rich and free resource \cite{yao2019tip,yao2020exploiting,yao2019towards,yao2016automatic}. For arbitrary categories, potential training data can easily be obtained from image search engines like Google or Bing. Therefore, it is intuitive to directly leverage web images to train fine-grained classification models. Unfortunately, due to the error index of image search engines, the precision of returned images is still unsatisfactory. For example, the average precision of the top 1,000 images for 18 categories from the Google Image Search Engine is merely 32\% \cite{schroff2011harvesting}. 

\begin{figure}[t]
\centering 
\includegraphics[width=0.45\textwidth]{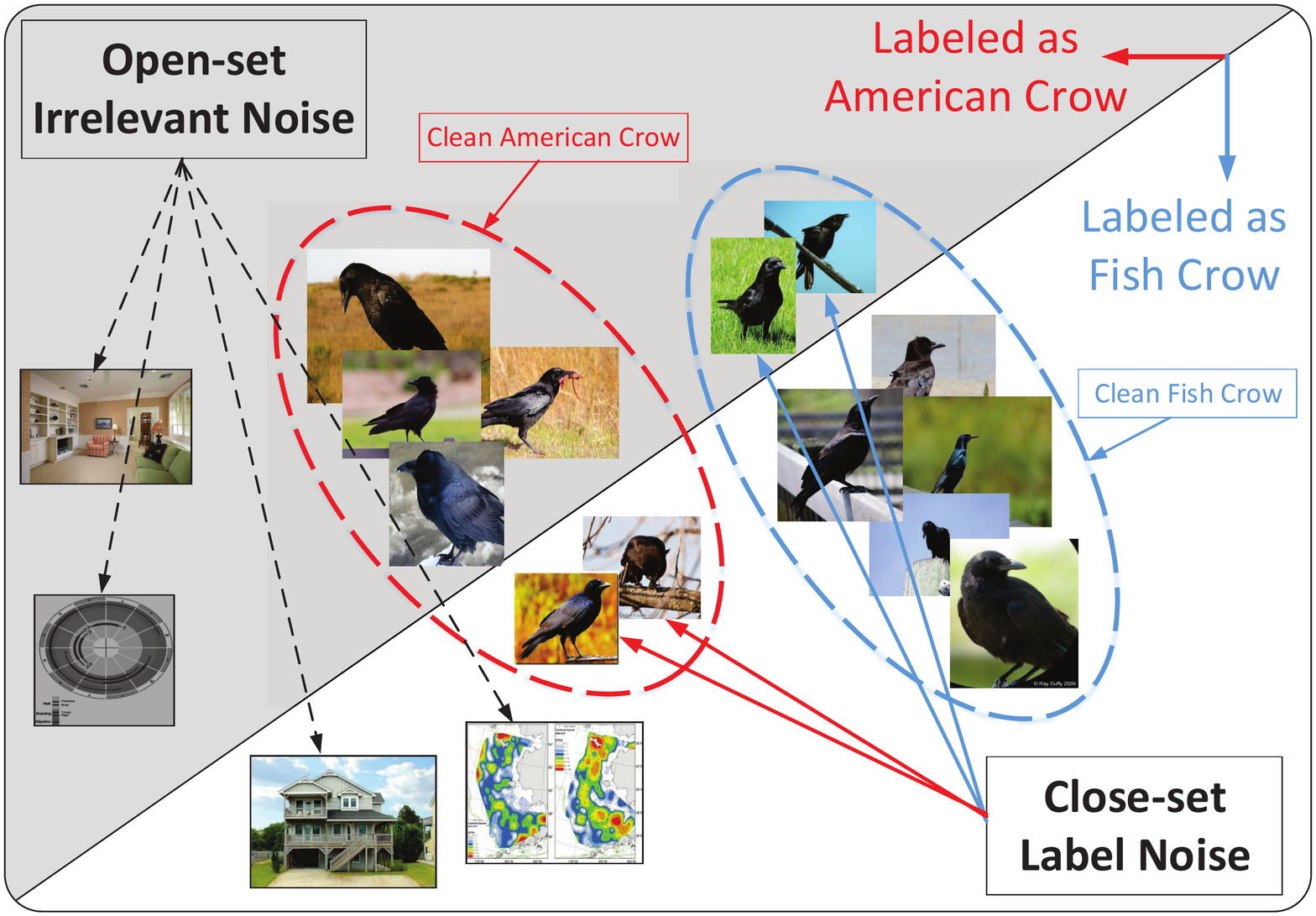} 
\caption{An illustration of the close-set and open-set noisy web images for a bird dataset. Specifically, noisy images in the close-set have their true labels in the dataset. Open-set contains irrelevant noisy images, whose labels are outside of the dataset.}
\label{fig1}
\end{figure}

As shown in Fig.~\ref{fig1}, noisy web images within fine-grained categories can be divided into two groups: close-set and open-set noise. Specifically, for noisy images in the close-set, their true labels can be found in the given dataset (\eg, the "fish crow" images are mistakenly labeled as "American crow" in Fig.~\ref{fig1}). The open-set, in contrast, consists of irrelevant noisy images, whose labels fall outside of the dataset (\eg, the images indicated by the black dotted arrows in Fig.~\ref{fig1}). As DNNs have a high capacity to fit noisy data \cite{arpit2017closer,zhang2016understanding}, training fine-grained recognition models directly with these noisy web images tends to result in poor performance.

To reduce the harmful influence of noise, some works have concentrated on estimating the noise transition matrix. For example, \cite{reed2014training} leveraged a bootstrapping loss and assigned a weight to the current prediction to compensate for the erroneous guidance of noisy samples. As an extension, \cite{goldberger2016training} introduced an additional softmax layer to estimate the label noise transition matrix. However, recovering the exact noise transition matrix is difficult. Alternatively, another line of works have chosen to focus on the sample section mechanism, which aims to separate clean samples from noise. The representative works in this area include Decoupling \cite{malach2017decoupling} and Co-teaching \cite{han2018co}. These works identify clean samples within a mini-batch and use them to update the networks. Nevertheless, they assume a close-set noisy label setting. 

Such a restricted assumption contradicts the fact that open-set scenarios are more common in practice. Furthermore, ignoring the existence of irrelevant noise makes the models less robust. To address this, with the increased popularity of web-supervised learning, \cite{wang2018iterative,liang2017enhancing} instead focused on open-set scenarios to tackle irrelevant noise. Unfortunately, neither was designed for fine-grained visual classification.

In addition to noisy images, web datasets tend to contain many hard examples. For instance, some web images are corrupted by text or watermarks, while others may contain a small object with a large background. These hard examples rarely exist in labeled training sets. Compared with easy examples, the hard examples found in web datasets carry less useful foreground information and more irrelevant background knowledge. The irrelevant information can misguide the model and consequently degrade its robustness. This can become problematic when a model is directly trained using a large number of hard examples. However, if the hard examples are correctly leveraged, the performance of the model can actually be improved, without degrading the robustness.

Therefore, we propose an approach that removes irrelevant noise from the training set, while simultaneously employing useful hard examples during training. Our work is motivated by the following observations: (1) Soft labels contain more information than one-hot labels \cite{hinton2015distilling}, especially for fine-grained visual classification, where subcategories share obvious similarities. (2) DNNs always memorize easy instances first, and gradually adapt to hard and noisy examples \cite{arpit2017closer,zhang2016understanding}. (3) Label smoothing prevents overfitting by encouraging models to be less confident \cite{szegedy2016rethinking}.

During training, our approach aims to identify irrelevant noisy samples and dynamically drop them. In addition, it effectively utilizes hard examples to increase the robustness of the model. Unlike most existing methods, which use the loss values to find noisy samples, our approach instead leverages the cross-entropy of the softmax probability across consecutive epochs which we refer to as probability cross-entropy from here on. Our proposed approach can make effective use of the information encoded in the soft labels and is able to measure any changes in network prediction. This is because open-set noisy samples are harder to fit than clean ones, so their predictions are unstable during training, resulting in a high probability of cross-entropy. Thus, irrelevant noisy samples can be distinguished from the useful training set and dropped by calculating the probability cross-entropy. In this way, the proposed approach can alleviate the harmful effect of open-set noise and achieve better performance. To guide the model to learn more efficiently from hard examples, normalization and label smoothing are utilized to boost the model training. Extensive experiments and ablation studies demonstrate that our approach outperforms state-of-the-art methods. The main contributions of this work can be summarized as follows:

\begin{enumerate}
	
\item We propose a novel approach to remove irrelevant noisy samples and utilize hard examples from web images to train an FGVC model. Unlike existing methods, our approach can dynamically increase the drop rate, enabling it to retain more instances during early epochs and increasingly drop noisy images before they are memorized.
	
\item Our proposed global sampling-based approach can effectively overcome the noise rate imbalance problem common in web images. Specifically, this problem occurs when the number of noisy images fluctuates among different batches, meaning that clean samples might be dropped in some mini-batches, while noisy samples are used for training in others.
	
\item Extensive experiments demonstrate that our approach outperforms state-of-the-art methods. Our learning paradigm delivers a new pipeline for fine-grained visual classification, which is more practical for real-world applications.
	
\end{enumerate} 

The rest of the paper is organized as follows: Section II discusses related works. We propose our framework and associated algorithms in Section III. The experimental evaluations and discussions are presented in Section IV. Section V concludes the paper.

\section{Related Work}

In this section, we review related works from three aspects: fine-grained visual classification, web-supervised learning, and normalization in neural networks.

\begin{figure*}[t]
\centering 
\includegraphics[width=0.99\textwidth]{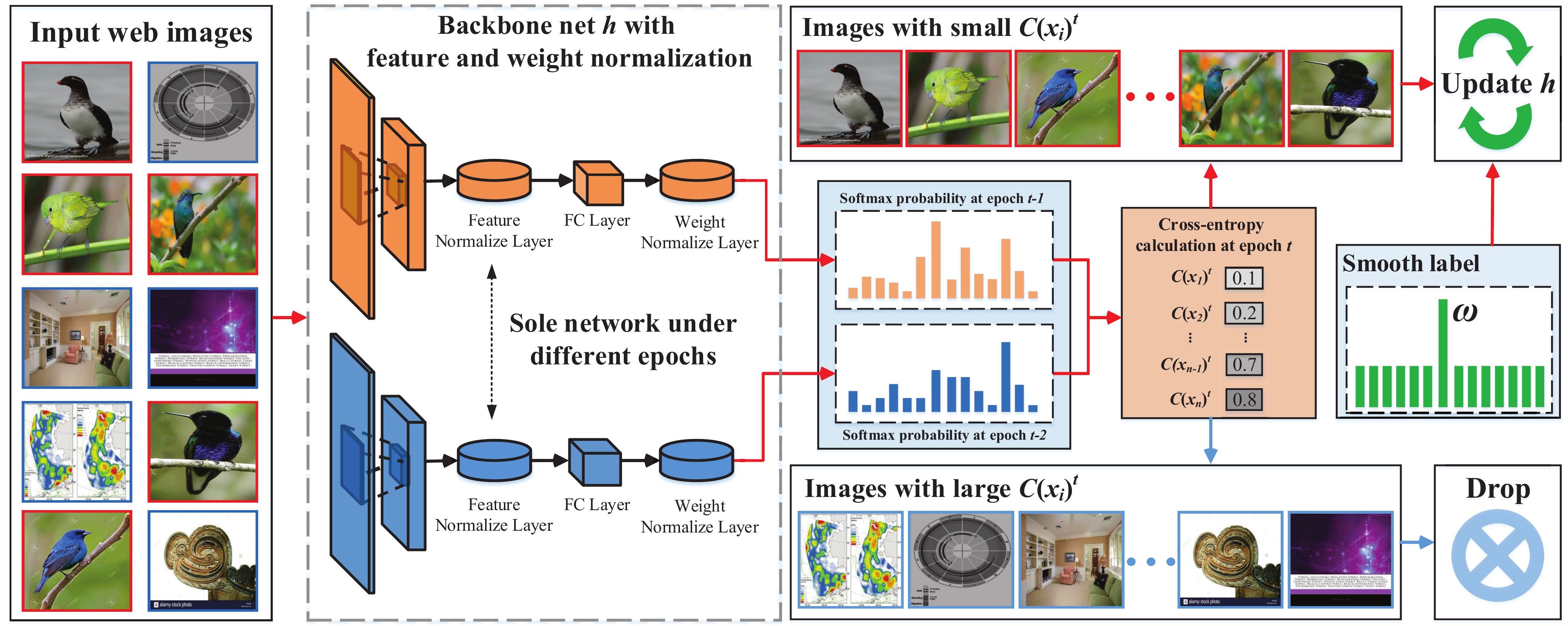} 
\caption{{The architecture of our model. For each input web image $ x_{i} $, our approach first obtains the softmax probability of epoch $ t-1 $, $ t-2 $ as $\textbf{\textit{p}}(x_{i})^{t-1}$ and $\textbf{\textit{p}}(x_{i})^{t-2}$, respectively. Then it computes the cross-entropy $C(x_{i})^{t}$ between $\textbf{\textit{p}}(x_{i})^{t-2}$ and $\textbf{\textit{p}}(x_{i})^{t-1}$ in epoch $ t^{} $. $C(x_{i})^{t}$ is leveraged to supervise the separation of useful and irrelevant noisy web images. To be specific, images with large $C(x_{i})^{t}$ are identified as irrelevant noisy images and then dropped during training. Those with small $C(x_{i})^{t}$ are regarded as useful images and utilized to further update the network $ \textbf{\textit{h}} $ with the smooth label, where the weight of the image label is $ \omega $.}}
\label{fig2}
\end{figure*}

\subsection{Fine-Grained Visual Classification}

The aim of fine-grained visual classification is to distinguish objects at a subordinate level. Taking into account the similarities between subcategories, early works typically trained the network to learn discriminative features using strong annotation cues, such as bounding boxes or part annotations \cite{huang2016part,yao2016coarse,lam2017fine,wei2018mask,zhang2014part}. Although they obtain promising results, these strongly supervised methods require heavy human annotation. To overcome this drawback, recent studies have focused on weakly supervised methods, which only require image-level labels \cite{fu2017look,lin2015bilinear,zheng2017learning,wang2018learning,ge2019weakly,zheng2019looking,chen2019destruction,korsch2019classification}. The state-of-the-art weakly supervised methods \cite{ge2019weakly,korsch2019classification} have shown competitive performance with strongly supervised methods. However, the label annotation still requires expert knowledge. This limits the dataset size. In addition, to mine discriminative features on small datasets, state-of-the-art weakly supervised methods tend to involve complex operations, such as part estimation \cite{korsch2019classification} or context encoding \cite{ge2019weakly}. To further improve the performance, several semi-supervised methods have managed to leverage widely available web data \cite{niu2018webly,cui2016fine,xu2016webly,niu2015visual,xiao2015learning,yao2020bridging,sun2020crssc,zhang2020data} for FGVC. For example, Niu \al \cite{niu2018webly} jointly utilized web data and well-labeled auxiliary data for FGVC, while Cui \al \cite{cui2016fine} leveraged an iterative approach for dataset bootstrapping and model training. Nevertheless, these semi-supervised methods still require human intervention. Different from semi-supervised methods, our approach is a pure web-supervised approach, which requires no human intervention.

\subsection{Web-Supervised Learning}

Training fine-grained recognition models with web images usually results in poor performance due to the presence of label noise and data bias \cite{zhang2020web}. Statistical learning has contributed significantly to solving this problem, especially in theoretical aspects \cite{ding2020approximate,yao2016domain,yao2019tmm,yao2018discovering}. In this work, we focus on deep learning based approaches. Roughly speaking, these works can be separated into four groups. The first group involves developing novel loss functions~\cite{reed2014training,pencil2019} to deal with label noise. The second group tries to estimate the noise transition matrix~\cite{patrini}. The third one applies attention mechanisms to alleviate noise and data bias~\cite{zhuang2017attend}. The last group attempts to clean the web data as a preprocessing step~\cite{malach2017decoupling,han2018co,mnet}. However, none of these works are specifically designed for fine-grained visual recognition.

\subsection{Normalization}

Normalization has been widely used in person recognition \cite{cosineloss} and face verification \cite{l2softmax,amsoftmax}. For example, \cite{l2softmax} utilized feature normalization to make the model pay more attention to blurry images. AM-softmax \cite{amsoftmax} further improved the effect of normalization by introducing an additive angular margin for the softmax loss. The additive angular margin makes the learned features more compact and therefore reduces the intra-class distance. Normalization significantly improves the performances of these methods in person recognition and face verification tasks. Inspired by these works, we utilize feature and weight normalization for the FGVC task.

\section{The Proposed Approach}

As shown in Fig.~\ref{fig2}, our approach consists of three main steps: normalization, denoising, and hard example utilization. We {provide} details of each step as follows.

\subsection{Normalization}

Following \cite{amsoftmax}, the feature $ f $ and weight $ W $ are normalized ($\left\|W_{j}\right\|=\left\|f_{i}\right\|=1 $) in the nonbias softmax loss:
\begin{equation}
	\mathcal{L}_{S}(x_{i},y_{i})=-\log\dfrac{\mathrm{e}^{W_{y_{i}}^{\mathrm{T}}f_{i}}}{\sum_{j=1}^{M}\mathrm{e}^{W_{j}^{\mathrm{T}}f_{i}}},
	\label{softmax}
\end{equation}
where $ x_{i} $ denotes the \textit{i}-th sample with label $ y_{i} $, $ f_{i} $ is the input feature of the last fully connected layer, $ W_{j} $ is the \textit{j}~-th column of the last fully connected layer, and $ \textit{M} $ is the number of categories. 
{Thus, we have 
	\begin{equation}
	W_{y_{i}}^{\mathrm{T}}f_{i} = \left\|W_{y_{i}}\right\| \left\|f_{i}\right\| \cos\theta_{y_{i}} = \cos\theta_{y_{i}}, 
	\end{equation}
	where $ \theta_{y_{i}} $ is the angle between vectors $ W_{y_{i}} $ and $ f_{i} $. After normalization ($\left\|W_{j}\right\|=\left\|f_{i}\right\|=1 $), the network outputs the cosine distance between $ W_{y_{i}} $ and $ f_{i} $. }
Then, the cosine values are scaled with a hyperparameter \textit{s} and the normalized loss function becomes:
\begin{equation}
	\mathcal{L}_{N}(x_{i},y_{i})=-\log\dfrac{\mathrm{e}^{s\cdot \cos\theta_{y_{i}}}}{\sum_{j=1}^{M}\mathrm{e}^{s\cdot \cos\theta_{j}}},
	\label{normalized_softmax}
\end{equation}
where $ \theta_{j} $ is the angle between vectors $ W_{j} $ and $ f_{i} $. The scaling factor \textit{s} is used to accelerate and stabilize the optimization. Wang \al \cite{Normface} pointed out that the network fails to converge without this scaling factor \textit{s}. Following \cite{amsoftmax,Normface}, the value of $ \textit{s} $ is set to 30. 

\begin{figure}[t]
\centering 
\includegraphics[width=0.49\textwidth]{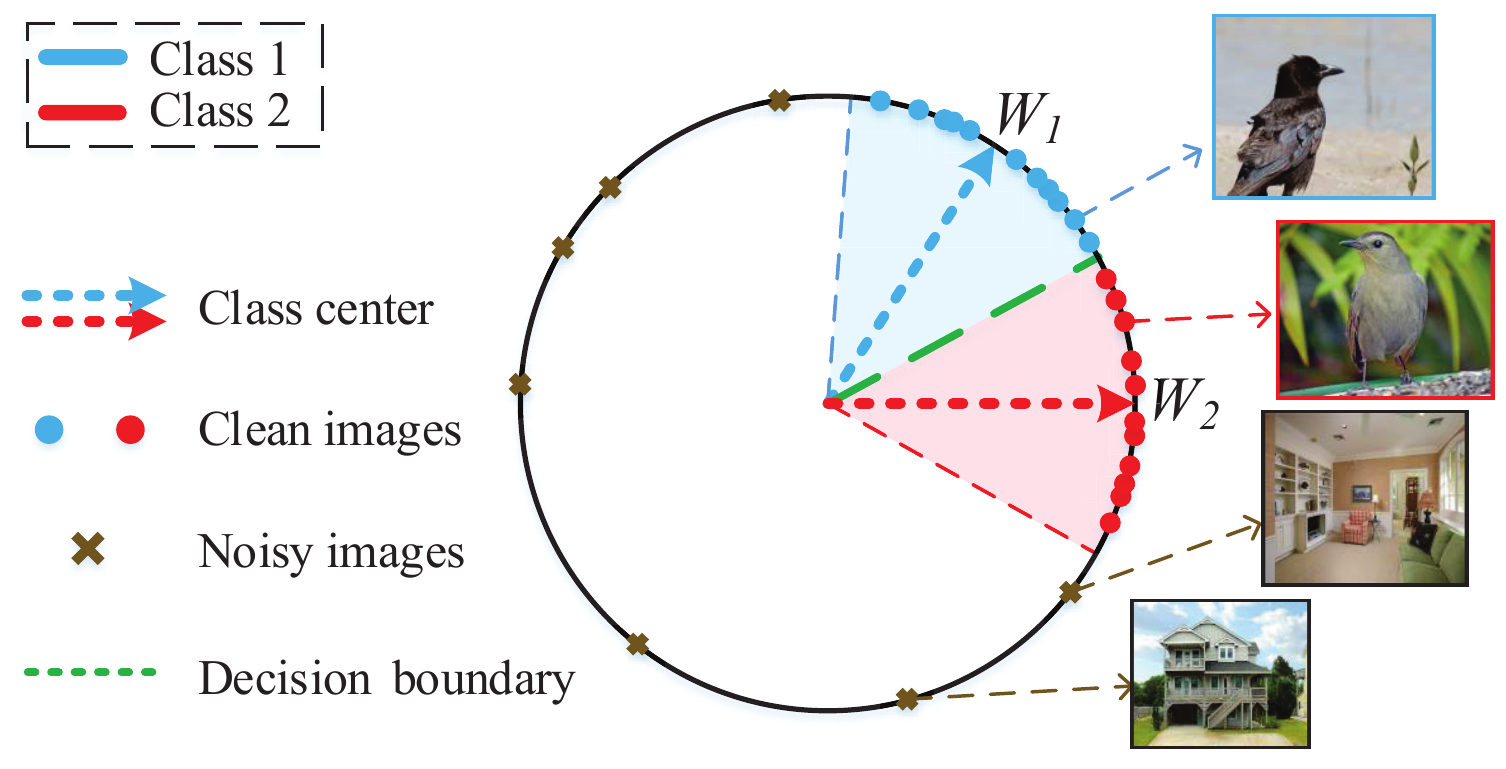} 
\caption{{After normalization, features are angularly distributed on a hypersphere, and the network learns the center of each class $ W_{j} $ during training.}}
\label{fig3-2}
\end{figure}

Fig.~\ref{fig3-2} shows the feature distributions after normalization. As can be seen, the features are angularly distributed on a hypersphere, because their $ L_{2} $-norms are scaled to 1. Accordingly, the last fully connected layer learns the center of each class $ W_{j} $ during training. Ranjan \al \cite{l2softmax} revealed that hard examples, especially blurry images, tend to have a lower $ L_{2} $-norm. After feature normalization, features with small norms will obtain a much higher gradient compared to features that have large norms \cite{amsoftmax}. As a consequence, the model will pay more attention to these hard examples during back-propagation. In this way, feature normalization enables the model to learn more effectively from hard examples. 

\subsection{Denoising}

Web-supervised learning is an open-set problem. Our approach first identifies and drops the irrelevant noisy images to purify the web training set.

\subsubsection{Noise Identification}

Memorization effects \cite{arpit2017closer,zhang2016understanding} indicate that deep neural networks always fit easy examples in the initial epochs, and then gradually adapt to hard and noisy examples. Note that, irrelevant noisy samples are totally different from clean ones in the training set, with diverse patterns. As a result, the model struggles to fit to them. The prediction results of irrelevant noisy samples thus change dramatically during training, especially at the early stages. Accordingly, we can identify noisy samples by measuring the changes in prediction.

Let $(x_{i}, y_{i})$ be a pair comprising sample $ x_{i} $ and its label $ y_{i} $, and $ \mathcal{D} = \{(x_{i}, y_{i})|1 \leq i \leq N\} $ be the noisy web training set. Assume that the neural network $ \textbf{\textit{h}} = (h_{1},...,h_{M} ) $ is trained to classify M classes. At each epoch $ t^{} $, we first utilize the network output logits $ \textbf{\textit{h}}(x_{i}) $ to compute the softmax probability $\textbf{\textit{p}}(x_{i})^{t} = (p_{1}(x_{i})^{t},...,p_{M}(x_{i})^{t})$ for each instance $x_{i}$ in the training set $\mathcal{D}$:
\begin{equation}
p_{j}(x_{i})^{t}=\dfrac{\mathrm{e}^{h_{j}(x_{i})}}{\sum_{s = 1}^M \mathrm{e}^{h_{s}(x_{i})}}.
\label{eq1}
\end{equation}

Then when epoch $ t^{} > $ 2, for each instance $x_{i}$, we compute the softmax probability cross-entropy $C(x_{i})^{t}$ between $\textbf{\textit{p}}(x_{i})^{t-2}$ and $\textbf{\textit{p}}(x_{i})^{t-1}$ through
\begin{equation}
C(x_{i})^{t}=-\sum_{j = 1}^M p_{j}(x_{i})^{t-1}\log p_{j}(x_{i})^{t-2}. 
\label{eq2}
\end{equation}

The probability cross-entropy $C(x_{i})^{t}$ reveals the changes in prediction. Since the predictions of irrelevant noisy samples change more rapidly than those of clean ones, they have higher probability cross-entropy values. For useful samples $x$ in the selected set $ \widehat{\mathcal{D}} $ and noisy samples $ \widetilde{x} $ in the dropped set $ (\mathcal{D} - \widehat{\mathcal{D}}) $, we have
\begin{equation}
\dfrac{1}{|\mathcal{D} - \widehat{\mathcal{D}}|}\sum_{\widetilde{x} \in (\mathcal{D} - \widehat{\mathcal{D}})}C(\widetilde{x}) > 	\dfrac{1}{|\widehat{\mathcal{D}}|}\sum_{x \in \widehat{\mathcal{D}}}C(x).
\end{equation}

Then, we can select samples that have a low probability cross-entropy $ C(x)^{t} $ as useful images and use them to update the network at each epoch $t_{}$. Different from existing methods that directly leverage cross-entropy, our method utilizes the probability, a soft label, to identify noisy samples. In this way, our approach can distinguish irrelevant noise from useful images in a web training set more efficiently than existing methods.

%
%
%
%

{
\begin{algorithm}[t]
	\SetAlgoLined
	\small
	\caption{Softly Update-Drop Training}
	\KwInput{Initialized network \textbf{\textit{h}}, training set $ \mathcal{D} $, {maximum drop rate} $\tau$, epoch $t_{k}$ and $t_{\max}$.}
	
	\For{$t = 1, 2, ..., t_{\max}$}{
		\For{\textup{each instance} $x_{i}$ \textup{in training set} $ \mathcal{D} $}{
			\If{$t > 2$}{\textbf{Compute} $C(x_{i})^{t}$ according to Eq.~(5)}
			
			\textbf{Compute} $\textbf{\textit{p}}(x_{i})^{t}$ according to Eq.~(4)	
		}
		
		\textbf{Update} $\textit{r}(t) $ according to Eq.~(7)
		
		\eIf{$t > 2$}{\textbf{Obtain} selected set $ \widehat{\mathcal{D}}^{t} $ according to Eq.~(8)}
		{\textbf{Set} $ \widehat{\mathcal{D}}^{t} = \mathcal{D} $}
		
		\textbf{Update} \textbf{\textit{h}} according to Eq.~(11)
	}
	\KwOutput{Updated network \textbf{\textit{h}}}
	\label{algorithm}
\end{algorithm}
}

\begin{table*}[t]
	\centering
	\renewcommand{\arraystretch}{1.1}
	\caption{ACA (\%) performances on three fine-grained benchmark datasets. BBox/Anno (\checkmark) indicates that human annotations are utilized during training. "Training set" shows whether the dataset is manually labeled (anno.) or collected from the web (web). iNat denotes the iNaturalist dataset.}
	\begin{tabular}{c|r|c|c|c|c|c|c}
		\hline
		\multirow{2}{*}{\textbf{Supervision}} & \multirow{2}{*}{\textbf{Method\:\:\:\:\:\:\:\:\:\:\:\:\:}} & \multirow{2}{*}{\textbf{Publication}} & \multirow{2}{*}{\textbf{BBox/Anno}} & \multirow{2}{*}{\textbf{Training Set}} & \multicolumn{3}{c}{\textbf{Datasets}} \\
		\cline{6-8} &                &                      &          &    & CUB200-2011  & FGVC-Aircraft & Cars-196\\		
		
		\hline
		\multirow{4}{*}{Strongly}
		
		&  Part-Stacked CNN \cite{huang2016part}	& CVPR 2016 & \checkmark  & anno.      &  76.60  &  - & - \\
		&  Coarse-to-fine \cite{yao2016coarse}	& TIP 2016  & \checkmark  & anno.      &  82.90  &  87.70 & - \\
		&  HSnet \cite{lam2017fine}				& CVPR 2017 & \checkmark  & anno.      &  87.50  &  - & 93.90 \\
		&  Mask-CNN \cite{wei2018mask} 		 	& PR 2018   & \checkmark  & anno.      &  85.70  &  - & - \\
		
		\hline
		\multirow{7}{*}{Weakly} 			
		
		&  Bilinear CNN \cite{lin2015bilinear}      & ICCV 2015 &     & anno.      &  84.10  & 83.90 & 91.30\\
		&  RA-CNN \cite{fu2017look}                 & CVPR 2017 &     & anno.      &  85.30  & -     & 92.50\\
		&  Multi-attention \cite{zheng2017learning} & ICCV 2017 &     & anno.      &  86.50  & 89.90 & 92.80\\
		&  Filter-bank \cite{wang2018learning}      & CVPR 2018 &     & anno.      &  86.70  & 92.00 & 93.80\\
		&  Parts Model \cite{ge2019weakly}		   & CVPR 2019 &     & anno.      &  90.40  & -		& -	   \\
		&  TASN \cite{zheng2019looking}	   		   & CVPR 2019 &     & anno.      &  89.10  & -		& 93.80\\
		&  DCL \cite{chen2019destruction}		   & CVPR 2019 &     & anno.      &  87.80  & 93.00	& 94.50\\
		
		\hline
		\multirow{4}{*}{Semi} 
		
		&  Xu \al \cite{xu2016webly}               & TPAMI 2018 & \checkmark  & anno.+web  & 84.60   & -	 & -\\
		&  Cui \al \cite{cui2016fine}              & CVPR 2016  & \checkmark  & anno.+web  & 80.70   & -	 & -\\
		&  Niu \al \cite{niu2018webly}             & CVPR 2018  &             & anno.+web  & 76.47   & -	 & -\\
		&  Cui \al \cite{cui2018large}             & CVPR 2018  &             & anno.+iNat & 89.29   & 90.70& 93.50\\
		
		\hline
		\multirow{8}{*}{Webly}

		&  WSDG \cite{niu2015visual}                  & CVPR 2015    &      & web &  70.61   & -	 & -\\
		&  Xiao \al \cite{xiao2015learning}           & CVPR 2015    &      & web &  70.92   & -	 & -\\
		&  Decoupling \cite{malach2017decoupling}     & NeurIPS 2017 &      & web &  70.56   & 75.97 & 75.00\\
		&  Co-teaching \cite{han2018co}               & NeurIPS 2018 &      & web &  73.85   & 72.76 & 73.10\\
		&  No-correction            				  & -			 &      & web &  66.57   & 64.33 & 67.42\\	
		&  Cross-entropy 							  & -    		 &      & web &  77.13   & 70.03 & 78.56\\
		&  Probability Cross-entropy       			  & -    		 &      & web &  77.22   & 72.88 & 78.71\\
		& \textbf{Ours}            			  		  & -            &      & web & \textbf{78.17}   &\textbf{77.95} &\textbf{83.50}
		\\
		\hline
	\end{tabular}
	\label{tab1}
\end{table*}

\subsubsection{Dynamic Drop Rate and Global Sampling}

Our approach selects samples from the whole training set and considers images with a low cross-entropy $C(x)^{t}$ as useful samples. By doing this, it can form a selected set $ \widehat{\mathcal{D}}^{t} $. Images with a high cross-entropy $C(x)^{t}$ are regarded as irrelevant noisy images and are not used for training. {The number of selected samples is controlled by a drop rate \begin{equation}\textit{r}(t) = \tau \cdot \min\{\frac{t}{t_{k}},1\}, \label{eq3}\end{equation} which is dynamically updated during training. Parameter $ t_{k} $ is the initial epoch number, which controls when the drop rate reaches the maximum value $ \tau $. In the early training stage ($ t \leq t_{k} $), $\textit{r}(t)$ rises steadily until it reaches the \textbf{maximum drop rate} $ \tau $. $ \widehat{\mathcal{D}}^{t} $ can be obtained by solving the following equation:
\begin{equation}
\widehat{\mathcal{D}}^{t} = \mathop{\arg\min}_{\mathcal{D'}:|\mathcal{D'}| \geq (1-\textit{r}(t))|\mathcal{D}|} \sum\nolimits_{x\ \in\ \mathcal{D'}} C(\textit{x})^{t}. 
\label{sample_selection}
\end{equation}
Eq.~\ref{sample_selection} indicates that our method updates $ \widehat{\mathcal{D}}^{t} $ at each epoch $ t^{} $ by selecting $ (1-\textit{r}(t)) \times 100\% $ samples with small $C(x)^{t}$ from the noisy training set $ \mathcal{D} $. Then, it leverages only the selected set $ \widehat{\mathcal{D}}^{t} $ to update the parameters of network \textbf{\textit{h}}.}

Similar to Co-teaching \cite{han2018co}, our approach utilizes a linearly increased drop rate $\textit{r}(t)$ in the early training epochs ($ t \leq t_{k} $). {As indicated by memorization effects \cite{arpit2017closer,zhang2016understanding,han2018co}, deep neural networks have the ability to filter out noisy instances using their loss values during the early training stage, but will eventually overfit to noisy samples as the number of epochs increases.} To leverage this property, our approach dynamically increases the drop rate $\textit{r}(t)$ and manages to retain more instances at early epochs and increasingly drop noisy images before they are memorized.

Existing methods tend to perform sample selection using mini-batches \cite{han2018co}. However, the number of noisy images $ N_{i} $ in a mini-batch \textit{i} follows a hypergeometric distribution. Given the noise rate $ R_{\mathcal{D}} $ of dataset $ \mathcal{D} $ and batch size $ N_{b} $, we have
\begin{equation}
N_{i}\sim H(|\mathcal{D}|,|\mathcal{D}| \cdot R_{\mathcal{D}},N_{b}).
\end{equation}
In this distribution, the number of noisy images $ N_{i} $ fluctuates among different batches, resulting in a noise rate imbalance problem. Specifically, some batches may have less noisy samples, while others have more. Thus, when the drop rate $\textit{r}(t)$ is fixed in each epoch, clean samples might have to be dropped in some mini-batches, while noisy samples used for training in others. Therefore, the sample selection in mini-batches is unstable and unreliable. To overcome the noise rate imbalance problem, our approach selects samples from the whole training set. By making the selection results more stable, better performance can be achieved.

\subsection{Hard Examples Utilization}

In addition to removing noisy samples, our approach is also able to properly utilize hard examples in the web training set. Since hard examples carry more irrelevant background information, they may misguide the model to learn irrelevant or invalid information. {This may result in a degradation of the model's generalization ability. If, on the other hand, the hard examples are used properly, they can instead improve the robustness of the model.} To address this contradiction, we propose to leverage label smoothing as it can prevent overfitting by making the model less confident about the predictions. Our approach assigns a label weight $ \omega $ for image label and $ \frac{1 - \omega}{M-1} $ for other categories. {Then the smooth loss functions are:
\begin{equation}
	\mathcal{L}_{Smooth}(x_{i},y_{i}) = \omega \cdot \mathcal{L}_{N}(x_{i},y_{i}) + \frac{1-\omega}{M-1} \cdot \sum_{j\neq y_{i}} \mathcal{L}_{N}(x_{i},j),
\end{equation}
and 
\begin{equation}
	{\mathcal{L}_{Final} = \dfrac{1}{|\widehat{\mathcal{D}}^{t}|}\sum_{x \in \widehat{\mathcal{D}}^{t}} \mathcal{L}_{Smooth}(x,y),}
	\label{final}
\end{equation}
where $ \textit{M} $ is the number of the categories, and $ j $ indicates each category except $ y_{i} $. Parameter $ \omega \in (0,1) $ controls the confidence of the prediction. A large $ \omega $ barely improves the generalization, while a small $ \omega $ may cause underfitting. In our experiments, we find that a moderate $ \omega $ can significantly improve the performance.} The detailed steps of our proposed approach are summarized in Algorithm~\ref{algorithm}.

\begin{figure*}[t]
\centering 
\includegraphics[width=0.99\textwidth]{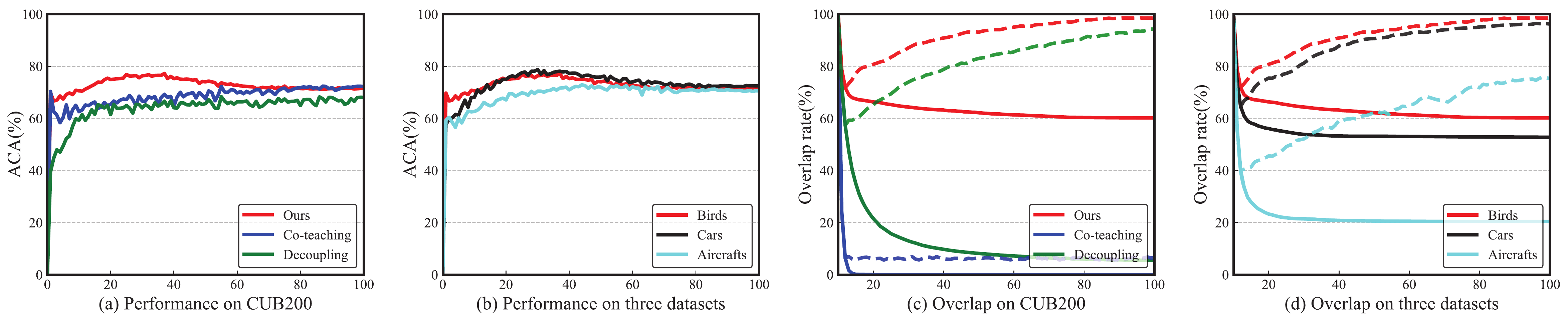}
\caption{Test accuracy and overlap rate vs. number of epochs. (a): Test accuracies of our approach, Co-teaching, and Decoupling on CUB200; (b): Test accuracies of our approach on three benchmark datasets; (c): Overlap rates of our approach, Co-teaching and Decoupling on CUB200; (d): Overlap rates of our approach on three benchmark datasets. The overlap rates of all previous epochs are plotted with solid lines, while the overlap rates of three contiguous epochs (\eg, epoch $ t_{i-2} $, $ t_{i-1} $ and $ t_{i} $) are plotted with dotted lines in (c) and (d).}
\label{fig3}
\end{figure*}

\section{Experiments}

\subsection{Datasets and Evaluation Metric}

Unfortunately, we cannot use web images as a validation/test set, because their labels are potentially incorrect. However, we can evaluate our approach on three commonly used fine-grained benchmark datasets, CUB200-2011 \cite{wah2011caltech}, FGVC-aircraft \cite{aircraft}, and Cars-196 \cite{car196}.

\textbf{CUB200-2011} covers 200 different subcategories of bird. It contains 11,788 images in total: 5,994 for training and 5794 for testing. For each subcategory, 30 images are selected for training and 11-30 for testing. Each image has detailed annotations: a subcategory label, a bounding box around the object, 15 part locations, and 312 binary attributes.

\textbf{FGVC-aircraft} consists of 10,000 images of 100 aircraft model variants: 6,667 images for training and 3,333 for testing. The aircraft in each image is annotated with a tight bounding box and a hierarchical airplane model label.

\textbf{Cars-196}: contains 196 subcategories of car, and includes 16,185 images: 8,144 for training and 8,041 for testing. For each subcategory, 24-84 images are selected for training and 24-83 for testing. Each image is annotated with a subcategory label and a bounding box of the object.

Average Classification Accuracy (\textbf{ACA}) is taken as the evaluation metric, which is widely used for evaluating the performance of fine-grained visual classification. 

\begin{figure}[t]
	\centering
	\includegraphics[width=0.48\textwidth]{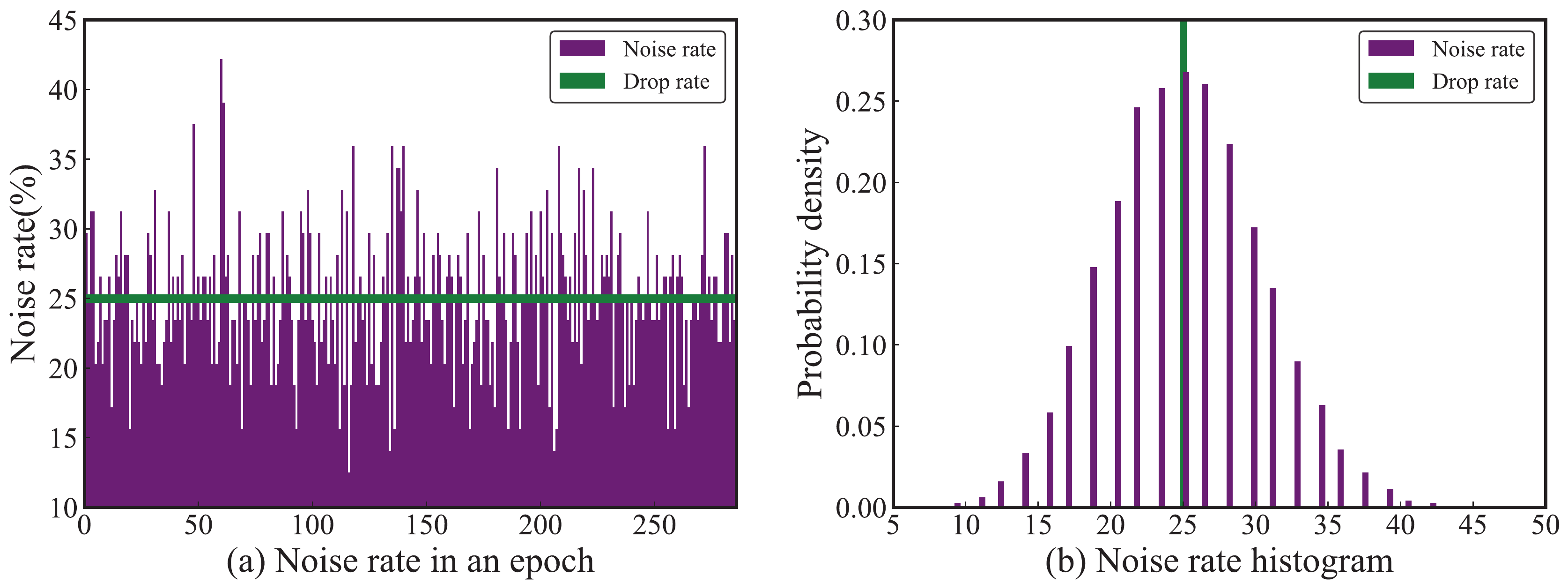}
	\caption{Noise rate of each mini-batch in an epoch (a) and noise rate histogram of all mini-batches (b).}
	\label{fig5}
\end{figure}

\subsection{Implementation Details} 

We directly utilize the web images collected in \cite{AAAI2020} and set these as the training set. We adopt the testing data from CUB200-2011, FGVC-aircraft, and Cars-196 as the test set for evaluation. Note that, we use two backbone networks, VGG-16 and ResNet-18. We select the maximum drop rate $ \tau $ from \{0.15, 0.2, 0.25, 0.3\}, epoch $t_{k}$ from \{5, 10, 15, 20\}, and label weight $ \omega $ from the range [0.1,0.9]. Based on the results of experiments, we ultimately set $ \tau=0.25 $ and $t_{k}=10$ as the default values for CUB200-2011 and Cars-196, and set $ \tau=0.20 $ and $t_{k}=10$ for the FGVC-Aircraft dataset. The label weight $ \omega $ is set to 0.6 for VGG-16 and 0.5 for ResNet-18.

For the VGG-16 backbone, we follow \cite{lin2015bilinear} and adopt a two-step training strategy. Specifically, we first freeze the convolutional layer parameters and only optimize the last fully connected layer. Then we optimize all layers in the previously learned model. In our experiments, we use an SGD optimizer with momentum = 0.9. The learning rate and number of epochs are set to 0.01 and 80 for both steps. The batch size is set to 64 and 32 for the first and second steps, respectively. We leverage a warm-up period of five epochs and then decay the learning rate through cosine annealing. For the ResNet-18 backbone, we utilize one-step training and set the batch size to 32. The other training settings are the same for the VGG-16 backbone.

\begin{figure}[t]
	\centering 
	\includegraphics[width=0.48\textwidth]{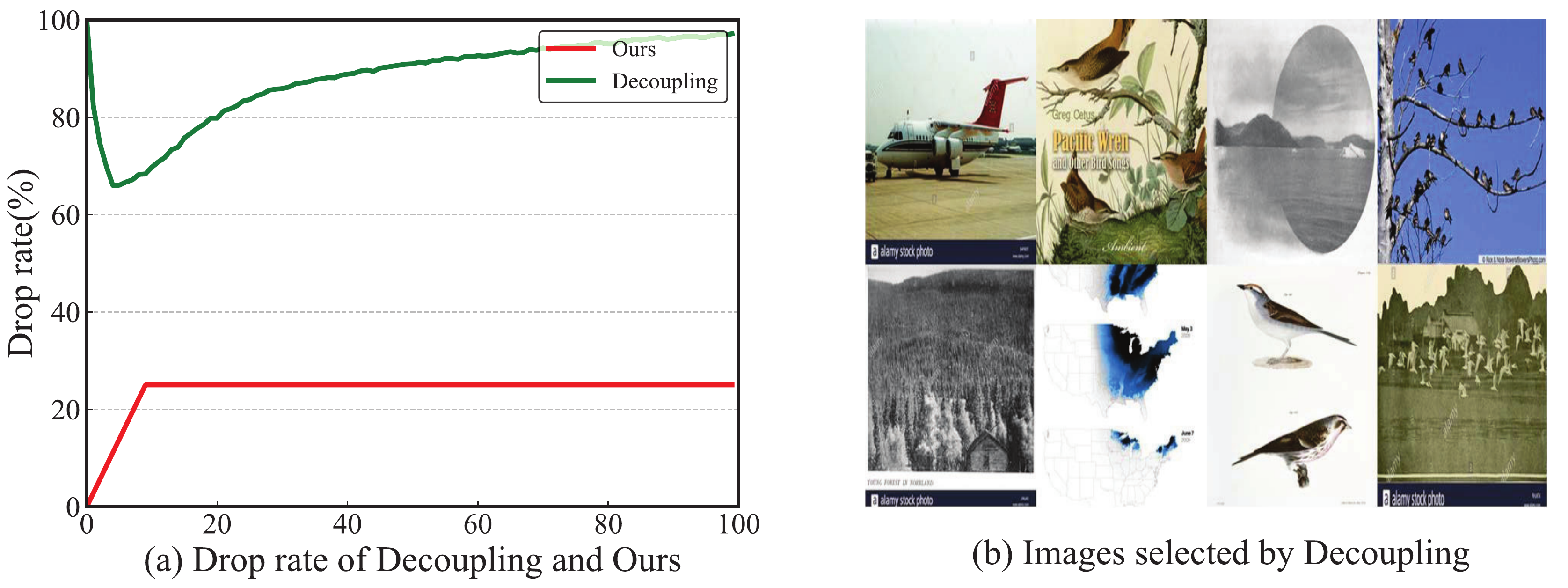}
	\caption{Drop rate of Decoupling (a) and sample images selected for training by Decoupling (b).}
	\label{fig8}
\end{figure}

\subsection{Baselines}

To illustrate the superiority of our approach, the following state-of-the-art methods are chosen as our baselines: \\
1) Strongly supervised fine-grained methods: Part-Stacked CNN \cite{huang2016part}, Coarse-to-fine \cite{yao2016coarse}, HSnet \cite{lam2017fine}, and Mask-CNN \cite{wei2018mask};
2) Weakly supervised fine-grained methods: Bilinear CNN \cite{lin2015bilinear}, RA-CNN \cite{fu2017look}, Filter-bank \cite{wang2018learning}, Multi-attention \cite{zheng2017learning}, Parts Model \cite{ge2019weakly}, TASN \cite{zheng2019looking}, and DCL \cite{chen2019destruction};
3) Semi-supervised fine-grained methods: \cite{xu2016webly}, \cite{niu2018webly}, \cite{cui2016fine}, and \cite{cui2018large};
4) Web-supervised methods: WSDG \cite{niu2015visual}, \cite{xiao2015learning}, Decoupling \cite{malach2017decoupling}, and Co-teaching \cite{han2018co}.
For Co-teaching and Decoupling, we replace the basic network with the same VGG-16 backbone network as ours and train the fine-grained models with the same web datasets. To be specific, we use all the same implementation settings except for the batch sizes, which are changed to 64 and 16 in the first and second steps, respectively. For Co-teaching, we set the maximum drop rate $ \tau=0.25 $ and epoch $t_{k}=10$. In addition, we train the VGG-16 backbone network without any correction (No-correction) and use cross-entropy to identify noise (Cross-entropy) for comparison. Experiments are conducted on one NVIDIA V100 GPU card.

\subsection{Performance of Denoising}

We first present the denoising performance of our approach, which does not utilize the normalization and label smoothing steps. 

\subsubsection{Experimental Results and Analysis}

Table~\ref{tab1} presents the fine-grained ACA results of various approaches on benchmark datasets. As can be seen in Table~\ref{tab1}, our proposed approach (Probability Cross-entropy) obtains significant improvements compared to other web-supervised methods on the CUB200-2011 and Cars-196 datasets. On the FGVC-Aircraft dataset, our approach achieves slightly better performance than Co-teaching.

Fig.~\ref{fig3} (a) presents the test accuracy vs. number of epochs for our approach, Decoupling, and Co-teaching on the CUB200 dataset. The memorization effect of networks can clearly observed in our approach, where the test accuracy quickly increases to a high level and then gradually decreases. In contrast, the test accuracies of Decoupling and Co-teaching rise slowly with obvious fluctuation, failing to reach a high level at the early stage of training. This is because our approach has a better sample selection ability, which enables it to reach a higher peak in much fewer epochs. Fig.~\ref{fig3} (b) shows the test accuracy vs. number of epochs on CUB200, FGVC-Aircrafts, and Cars-196. By observing Fig.~\ref{fig3} (b), a similar trend to that discussed above can be observed. 

\begin{figure}[t]
	\centering 
	\includegraphics[width=0.48\textwidth]{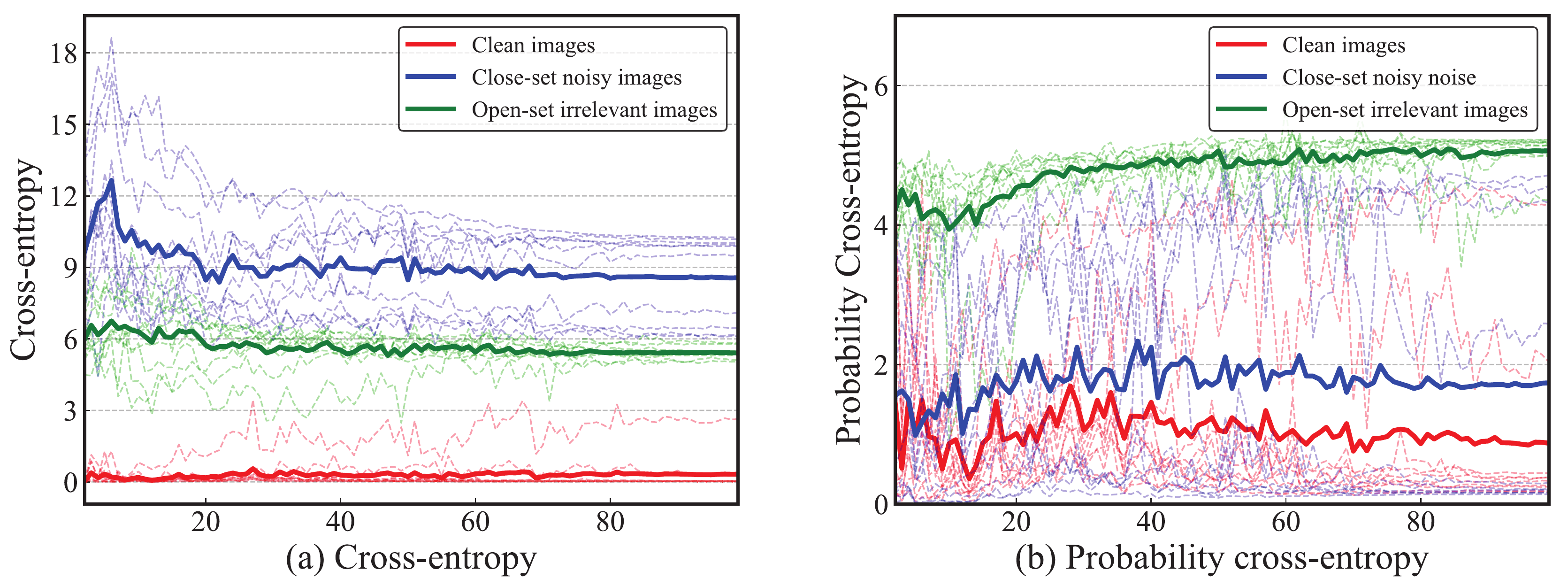}
	\caption{Cross-entropy (a) and Probability cross-entropy (b) of clean images, close-set noisy images and open-set irrelevant images. The value of each image is plotted in dotted line and the average value is plotted in solid line.}
	\label{fig4}
\end{figure}

To further demonstrate the sample selection ability of our approach, we record the selection result of each epoch during training for epoch $ t > t_{k} $ (we set $t_{k}$=10) and compute the \textit{overlap rate} of selected noisy samples. {We define the instance that is identified as a noisy sample in contiguous epochs (\eg \ epoch $ t_{i-2} $, $ t_{i-1} $ and $ t_{i} $, or all epochs) as an overlapped image. Given the number of overlapped noisy images $ N_{o} $, total number of training samples $ N $, and drop rate $\textit{r}(t)$, the overlap rate $ \textit{O} $ is computed by $ \textit{O}=\frac{N_{o}}{N\cdot\textit{r}(t)} $, which indicates the selection stability. Specifically, a higher overlap rate indicates stabler selection results. In contrast, if the selection results in different epochs are diverse, the overlap rate will be low.}

Fig.~\ref{fig3}~(c) shows the results of our approach, Co-teaching, and Decoupling on the CUB200 dataset. Fig.~\ref{fig3}~(d) provides the results of our approach on all three datasets.
Co-teaching leverages the same drop rate as ours. However, as the number of epochs increases, the overlap rate of all epochs in Co-teaching decreases to 0 rapidly, while our approach maintains a roughly stable number after a small drop (Fig.~\ref{fig3} (c)). Similarly, the overlap of contiguous epochs in Co-teaching remains small (around 10\%), while our approach clearly contains more overlap, which rises steadily as the number of epochs increases. This means that our approach maintains a stable selection result, which becomes more stable as the training continues. This improvement can be attributed to our global selection strategy. Specifically, Co-teaching performs sample selection in a mini-batch, where it cannot tackle the noise rate imbalance problem. Thus its selection result is unstable and changes rapidly during training. This further causes the network to learn from noisy images. By overcoming this drawback, our approach has better sample selection consistency and performance. From Fig.~\ref{fig3} (d), we can observe that our approach maintains stable sample selection results on all three datasets, especially on CUB200 and Cars-196.

\begin{figure}[t]
\centering
\includegraphics[width=0.48\textwidth]{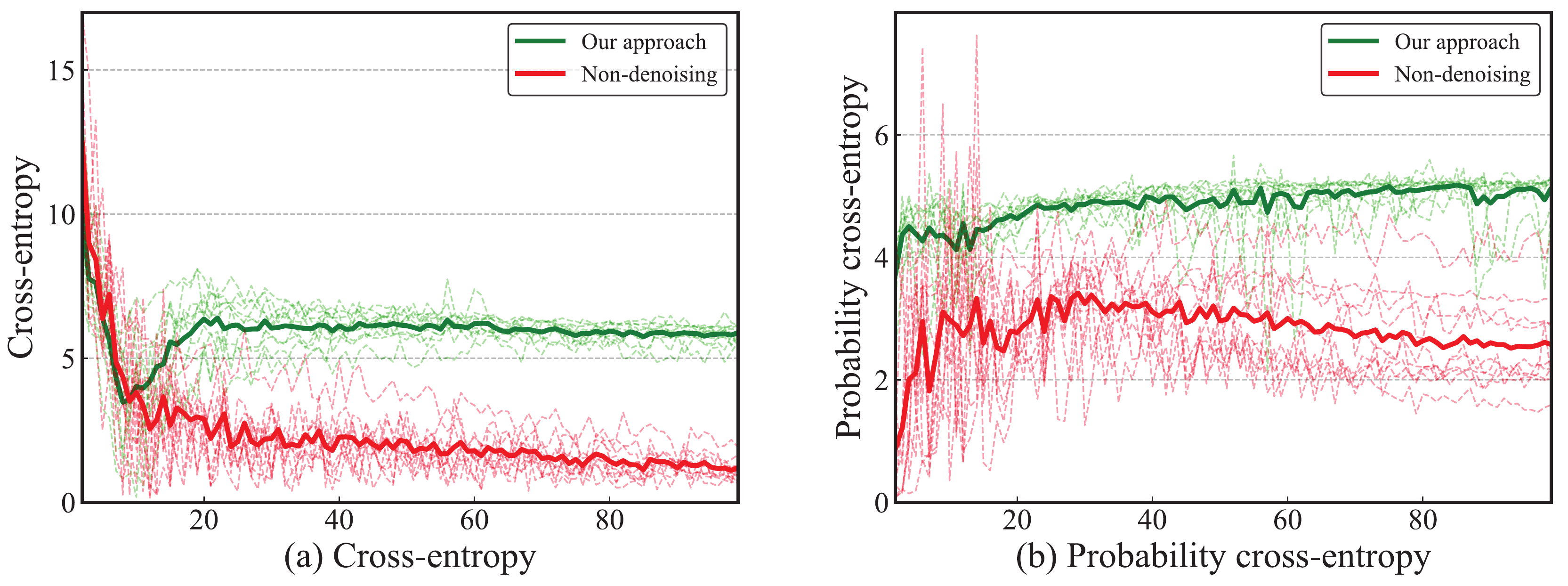}
\caption{Cross-entropy (a) and Probability cross-entropy (b) of irrelevant noisy images in our approach and non-denoising training. The value of each image is plotted in dotted line and the average value is plotted in the solid line.}
\label{fig6}
\end{figure}

\begin{figure}[b]
	\centering 
	\includegraphics[width=0.30\textwidth]{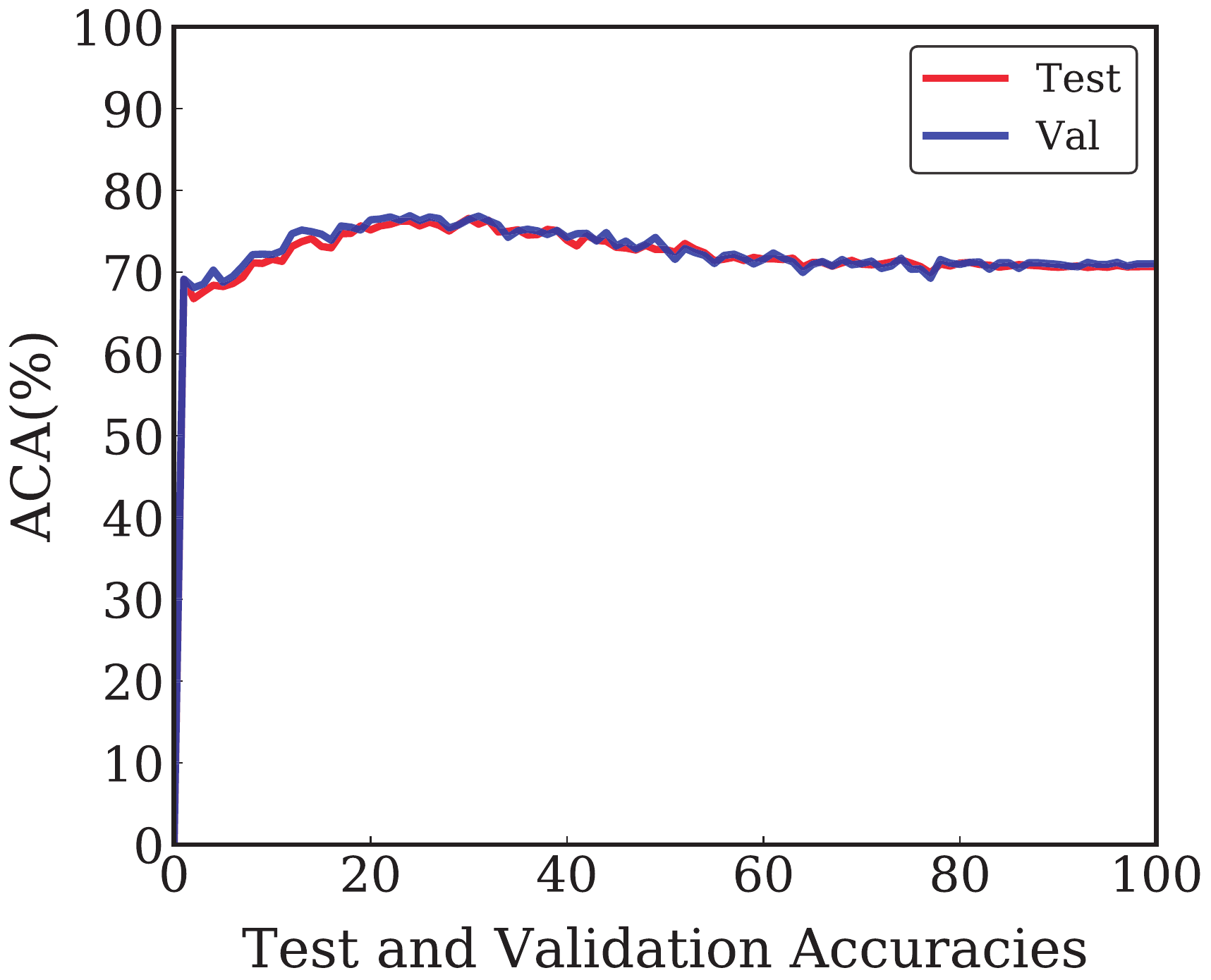} 
	\caption{Test and Validation Accuracies on CUB200.}
	\label{validation}
\end{figure}

\begin{table*}
	\centering
	\caption{ACA(\%) performances of different backbones, dataset sizes, and frameworks.}
	\begin{minipage}[t]{0.32\textwidth}
		\centering
		\renewcommand{\arraystretch}{0.95}
		\begin{tabular}{p{2.5cm}<{\centering}|p{2cm}<{\centering}} 
			\hline 
			\textbf{Backbones}	 &  \textbf{ACA (\%)}        \\
			\hline			
			VGG-16                      	 &  77.22                    \\ 
			VGG-19                      	 &  75.87                   \\
			ResNet-34                   	 &  74.99                    \\
			\hline
		\end{tabular}
	\end{minipage}
	\begin{minipage}[t]{0.32\textwidth}
		\centering
		\renewcommand{\arraystretch}{0.95}
		\begin{tabular}{p{2.5cm}<{\centering}|p{2cm}<{\centering}} 
			\hline
			\textbf{Dataset Sizes} &  \textbf{ACA (\%)}        \\
			\hline
			50                         		 &  71.87  					 \\ 
			75                        	 	 &  74.85				 \\ 
			100                         	 &  77.22 					 \\  
			\hline
		\end{tabular}
	\end{minipage}
	\begin{minipage}[t]{0.32\textwidth}
		\centering
		\renewcommand{\arraystretch}{0.95}
		\begin{tabular}{p{2.5cm}<{\centering}|p{2cm}<{\centering}} 
			\hline
			\textbf{Frameworks}	 &  \textbf{ACA (\%)}        \\
			\hline
			Co-teaching                      &  75.46                    \\ 
			Peer networks                    &  76.30                    \\	
			Single network                   &  77.22                    \\
			\hline
		\end{tabular}
	\end{minipage}
	\label{tab2}
\end{table*}

\begin{figure*}[t]
	\centering 
	\includegraphics[width=0.99\textwidth]{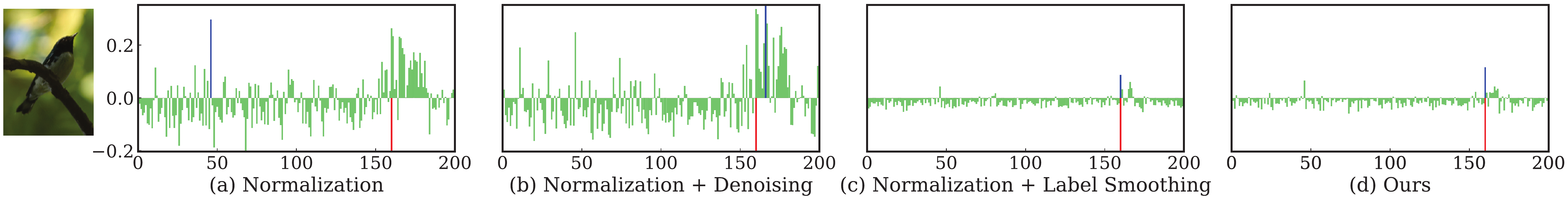}
	\caption{The logits of four models: (a) only normalization; (b) normalization and denoising; (c) normalization and label smoothing; (d) our approach. The red and blue bars indicate the true label and prediction of the test image, respectively.}
	\label{fig11}
\end{figure*}

\begin{table}[b]
	\centering
	\renewcommand{\arraystretch}{1.1}
	\caption{Training and testing ACA results (\%) of global and mini-batch selection mechanism.}
	\begin{tabular}{c|c|c|c} 
		\hline 
		\textbf{Backbone} & \textbf{Selection} & \textbf{Training (\%)} & \textbf{Testing (\%)}   \\
		\hline
		\multirow{2}{*}{VGG-16}	
		& Global 			&     76.53   & \textbf{77.22}   \\ 	
		& Mini-batch		&  	  83.21   & 75.66   \\ 		
		\hline
	\end{tabular}
	\label{tab3}
\end{table}

\subsubsection{Validation Set and Early Stop}

Given that the training set is noisy, we cannot simply separate a validation set out of the noisy web images. However, we can randomly choose 2,000 images from the CUB200 training set as a validation set. Specifically, we randomly choose 2,000 images from the CUB200 original training set as a validation set. The size of validation set is about one-third of the test set. We then perform the model selection based on the validation set. Finally, we use the selected model to perform the testing on the test set. We plot the accuracy vs. epoch curves by using the validation and test set, respectively. The experimental results are presented in Fig.~\ref{validation}. By observing Fig.~\ref{validation}, we can notice that the performance of our method increases firstly and then decreases. The explanation is that deep neural networks have memorization effects \cite{zhang2016understanding}. Deep neural networks will first learn the clean and easy patterns in the initial epochs. However, as the number of epochs increases, deep neural networks will eventually overfit on noisy labels. Therefore, a validation set for early stop and model selection is essential.   

\subsubsection{Noise Rate Imbalance}

We investigate the noise rate imbalance problem in mini-batches with noisy bird training images. We record the number of dropped images during training for each mini-batch. Assuming that the dropped images are noisy samples, we can compute the noise rate $ R_{i} $ of mini-batch \textit{i}. Given the number of dropped images $ N_{i} $ of mini-batch \textit{i} and batch size $ N_{b} $, we can calculate $ R_{i} $ by $R_{i}=\frac{N_{i}}{N_{b}}$. Fig.~\ref{fig5} (a) presents the noise rate of each mini-batch in a randomly selected epoch. It ranges from 12\% to 42\% with obvious fluctuation around the drop rate. To illustrate the distribution of noise rate $ R $, we record the noise rates of all mini-batches (more than 25,000) during training and plot the histogram of $ R $ in Fig.~\ref{fig5} (b). By observing Fig.~\ref{fig5} (b), we can notice that $ R $ follows a Gaussian distribution, ranging from 6\% to 48\%. Fig.~\ref{fig5} (a) explicitly shows the noise rate imbalance in mini-batches.
 
Decoupling selects samples with different predictions from two peer networks to update the model. It is a global selection method and does not have the noise rate imbalance problem, so it provides a stable selection result. As shown in Fig.~\ref{fig3} (c), the overlap rate of three contiguous epochs in Decoupling is close to our approach. However, the drop rate of Decoupling is too high. We record the number of dropped images and compute the drop rate of Decoupling in each epoch. The results are shown in Fig.~\ref{fig8} (a). From Fig.~\ref{fig8} (a), we can observe that the drop rate of Decoupling is always larger than 60\% and it climbs to nearly 100\% as training continues.

The extremely high drop rate demonstrates that Decoupling unable to make full use of the clean samples. Besides, since irrelevant noise is hard to fit to, the peer networks have a high probability of producing different predictions for these samples. Then, when the samples are used for training, they misguide the networks. Fig.~\ref{fig8} (b) visualizes some overlapped images which are used for training in Decoupling. These images are irrelevant noisy samples, indicating that Decoupling is not capable of tackling this irrelevant noise.

Moreover, we compare the performances of global and mini-batch selection in Table~\ref{tab3}. From Table~\ref{tab3}, we can observe that mini-batch selection has a higher training accuracy but lower testing accuracy than global selection. This suggests that mini-batch selection learns more training samples, some of which are noisy. It is then misguided by the noisy samples and achieves lower performance. {Performing sample selection in a mini-batch with a fixed drop rate is thus unable to tackle the noise rate imbalance problem. In contrast, our proposed approach, which leverages global sample selection, can mitigate this problem and achieve better performance.}

\begin{table*}[t]
	\begin{minipage}{0.5\linewidth}
		\renewcommand{\arraystretch}{1.1}
		\centering
		\caption{ACA (\%) performances of data augmentation. Anno. and web denotes the dataset is manually labeled and collected from the web, respectively.}
		\begin{tabular}{c|c|c|c} 
			\hline 
			\textbf{Models} & \textbf{Backbone} & \textbf{Training Set} & \textbf{Performance}   \\
			\hline
			\multirow{6}{*}{Ours}
			& ResNet-18 &     anno.         	& 81.53         \\ 	
			& ResNet-18 &  	  anno.+web         & \textbf{86.19}         \\ 	
			& VGG-16 	&     anno.         	& 82.79     \\ 	
			& VGG-16	&  	  anno.+web         & \textbf{86.26}         \\ 		
			& ResNet-50 &	  anno.				& 83.48			\\	
			& ResNet-50 &	  anno.+web			& \textbf{87.57}			\\
			\hline
		\end{tabular}
		\label{tab5}
	\end{minipage}
	\hspace{0.2cm}
	\begin{minipage}{0.5\linewidth} 
		\centering
		\renewcommand{\arraystretch}{1.1}
		\caption{ACA (\%) performance and improvement of VGG-16, ResNet-18 and ResNet-50. Baseline denotes the original backbone network.}
		\begin{tabular}{c|c|c|c} 
			\hline 
			\textbf{Backbone} & \textbf{Method} & \textbf{Performance} & \textbf{Improvement}   \\
			\hline
			\multirow{2}{*}{VGG-16}	
			& Baseline 	&     66.57         	&  
			\multirow{2}{*}{11.6} \\
			& Ours	    &  	  78.17          &  \\
			\hline
			\multirow{2}{*}{ResNet-18}			
			& Baseline &      68.59        	& 
			\multirow{2}{*}{8.58} \\ 
			& Ours     &  	  77.17         & \\
			\hline
			\multirow{2}{*}{ResNet-50}	 
			& Baseline &	  73.01				& 
			\multirow{2}{*}{7.38} \\ 
			& Ours     &	80.39  			& \\
			\hline
		\end{tabular}
		\label{tab6}
	\end{minipage}
\end{table*}

\subsubsection{Probability Cross-Entropy and Cross-Entropy}

In this experiment, we compare the performance of probability cross-entropy and cross-entropy in identifying noise on noisy bird training images. We first save the models produced each epoch during training. Then we leverage them to identify clean images, close-set noisy images, and open-set irrelevant images (30 images in total, 10 of each kind). We record their cross-entropy as well as probability cross-entropy. The experimental results are shown in Fig.~\ref{fig4}.
By observing Fig.~\ref{fig4} (b), we can notice that the probability cross-entropy of open-set irrelevant images is much larger than that of close-set noisy images and clean data. Compared with clean images, both close-set noisy images and open-set irrelevant images have a larger loss.    
{From Fig.~\ref{fig4} (a) and (b), we can conclude that selecting samples with cross-entropy cannot distinguish close-set and open-set noise}. Nevertheless, leveraging our proposed probability cross-entropy to identify open-set irrelevant images is reliable. {We also compare the performances when identifying noise by probability cross-entropy and cross-entropy in Table.~\ref{tab1}.} As demonstrated in Table.~\ref{tab1}, probability cross-entropy shows a slightly better performance on each dataset. One possible explanation is that some hard examples are regarded as noise by cross-entropy, because they tend to have larger cross-entropy values during training.

\subsubsection{Effectiveness of Denoising}

To explain the effectiveness of denoising proposed in our approach, we train the network using original web images without any denoising and record the probability cross-entropy and cross-entropy of 10 irrelevant noisy images during training. We also train the network using web images selected by our proposed approach and compare the results in Fig.~\ref{fig6}.
From Fig.~\ref{fig6} (b), we can find that the probability cross-entropy is large and declines extremely slowly during training, meaning that using probability cross-entropy can identify irrelevant noise during training and learn robust models. From Fig.~\ref{fig6} (a), we also notice that the cross-entropy in the non-denoising method gradually drops during training. In contrast, the cross-entropy in our approach drops slightly at first and then climbs to a roughly constant value. The explanation is that our approach has the ability to drop irrelevant noisy images before the network fits them.

\subsubsection{Influence of Different Backbones}

To investigate the influence of different CNN architectures in the denoising model, we replace VGG-16 with VGG-19 and ResNet-34. As shown in Table~\ref{tab2} (left), these three backbone networks achieve similar performances on CUB200. The performances of VGG-19 and ResNet-34 are slightly worse than that of VGG-16. One possible explanation is that we use the same coefficient settings for these backbones, which is best for VGG-16.

\subsubsection{Influence of Different Dataset Sizes}

We investigate the impact of data scale by changing the number of web images used for each category on CUB200. Specifically, we collect 50, 75, and 100 images from the web for each category.
As shown in Table~\ref{tab2} (middle), in general, the ACA performance improves steadily when using more training images. {Therefore, web-supervised learning is a promising research direction as it allows large-scale datasets to easily be built.}

\subsubsection{Influence of Multiple Networks}

We conduct two experiments to study whether using multiple networks can improve performance. In the first experiment, we combine our approach with the Co-teaching framework, letting the two networks select samples for each other. {In the second experiment, we use two peer networks and leverage their outputs to compute the probability cross-entropy.} {Given the softmax probabilities $\textbf{\textit{p}}(x_{i})^{t-1} $ and $\textbf{\textit{q}}(x_{i})^{t-1} $ of two peer networks,} the cross-entropy $ C(x_{i})^{t} $ is computed by
$C(x_{i})^{t}=-\sum_{j = 1}^N p_{j}(x_{i})^{t-1}\log q_{j}(x_{i})^{t-1}$.
The results are demonstrated in Table~\ref{tab2} (right). Both frameworks show worse performance than our proposed approach, which only utilizes a single network. Compared with methods that need two networks, our approach is lighter and more efficient.

{
\subsubsection{Summary of Denoising}

In our denoising experiments, we first illustrate that our denoising method outperforms other web-supervised methods in both performance and sample selection ability. Then we demonstrate that our global selection strategy can overcome the noise rate imbalance problem. Next, we compare our probability cross-entropy with cross-entropy, and illustrate that utilizing probability cross-entropy can identify open-set irrelevant images reliably. Finally, we investigate the influence of different backbones, dataset sizes and multi-networks. 
}

\subsection{Performances of Normalization and Label Smoothing}

In this subsection, we demonstrate the improvements provided by the normalization and label smoothing in our proposed full framework. The experimental results are shown in Table~\ref{tab1}. By observing Table~\ref{tab1}, we can notice that the normalization and label smoothing (full) remarkably improve the ACA performance on three datasets compared with only dropping irrelevant noisy images. The improvements are 0.95\%, 5.07\% and 4.79\% on CUB200-2011, FGVC-aircraft, and Cars-196 datasets, respectively. The explanation is that utilizing hard examples can effectively improve the robustness of the model.

\begin{figure}[b]
	\centering 
	\includegraphics[width=0.44\textwidth]{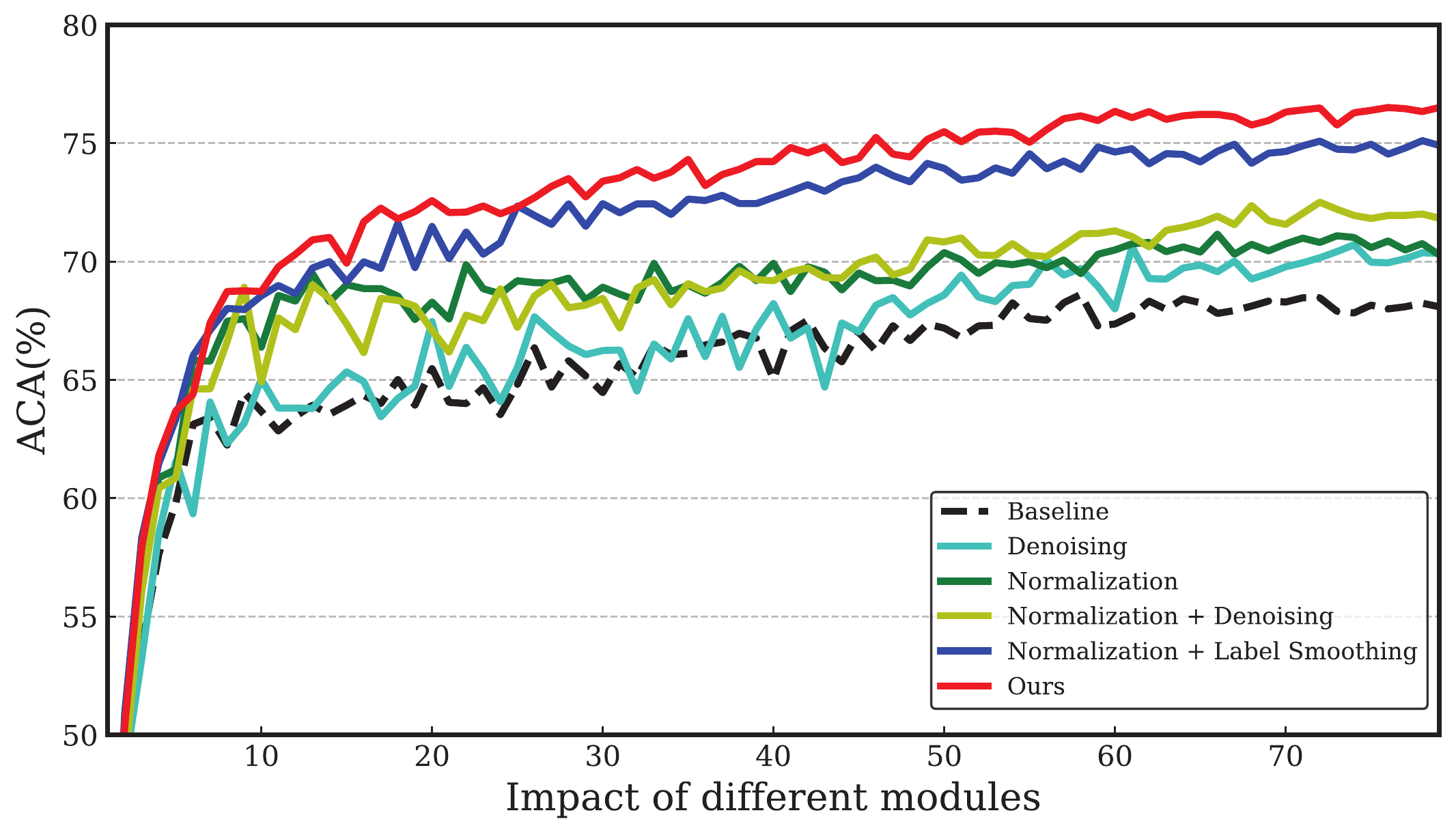} 
	\caption{Test accuracy vs. number of epochs for each module.}
	\label{fig14}
\end{figure}

\subsubsection{Influence of Different Modules}

Fig.~\ref{fig14} illustrates the impact of normalization, denoising, and label smoothing. This experiment is conducted with a ResNet-18 backbone because the impact of each module is more distinct on the ResNet-18 than VGG-16. To better demonstrate the improvements, we train a ResNet-18 model without any modules as the baseline for comparison. From Fig.~\ref{fig14}, we can observe that the baseline is outperformed by simply utilizing the normalization. On the basis of normalization, dropping noisy images or using label smoothing further improves the performance. Specifically, label smoothing contributes more to the improvement. Compared with using each module individually, our method shows much better performance as denoising and utilizing hard web images significantly improve the performance and training efficiency.
To further explore the effectiveness of denoising and label smoothing, we visualize the logits $ \textit{\textbf{h}}(x_{i}) $ of four models: only normalization; normalization and denoising; normalization and label smoothing, and our full approach. The results are shown in Fig.~\ref{fig11}. 

\subsubsection{Influence of Data Augmentation}

To demonstrate the advantage of leveraging web images, we train our model on the CUB200 dataset with our web images as data augmentation. To be specific, we set $ \omega $ = 0.8 and $ \omega $ = 0.5 for images in labeled and web datasets, respectively. {Images in the labeled dataset have less background information and deserve a higher label weight $ \omega $.} In addition, we train models on the labeled dataset for comparison. The results are given in Table~\ref{tab5}. From Table~\ref{tab5}, we can observe that leveraging web images remarkably improves the performance across different backbone networks. The improvements on VGG-16, ResNet-18 and ResNet-50 are 3.47\%, 4.66\% and 4.09\%, respectively. These results also indicate that leveraging web images is an effective way to improve the robustness of the model.

\subsubsection{Influence of Different Backbones}

To investigate the influences of different backbones in the whole framework, we conduct experiments on the CUB200 dataset with different backbone networks. The experimental results are demonstrated in Table~\ref{tab6}. By observing Table~\ref{tab6}, we can notice that our approach shows significant improvements across different backbone networks. The experimental results also indicate that our proposed approach is a robust and effective web-supervised method.

\subsubsection{Visualization}

\begin{figure}[t]
	\centering 
	\includegraphics[width=0.48\textwidth]{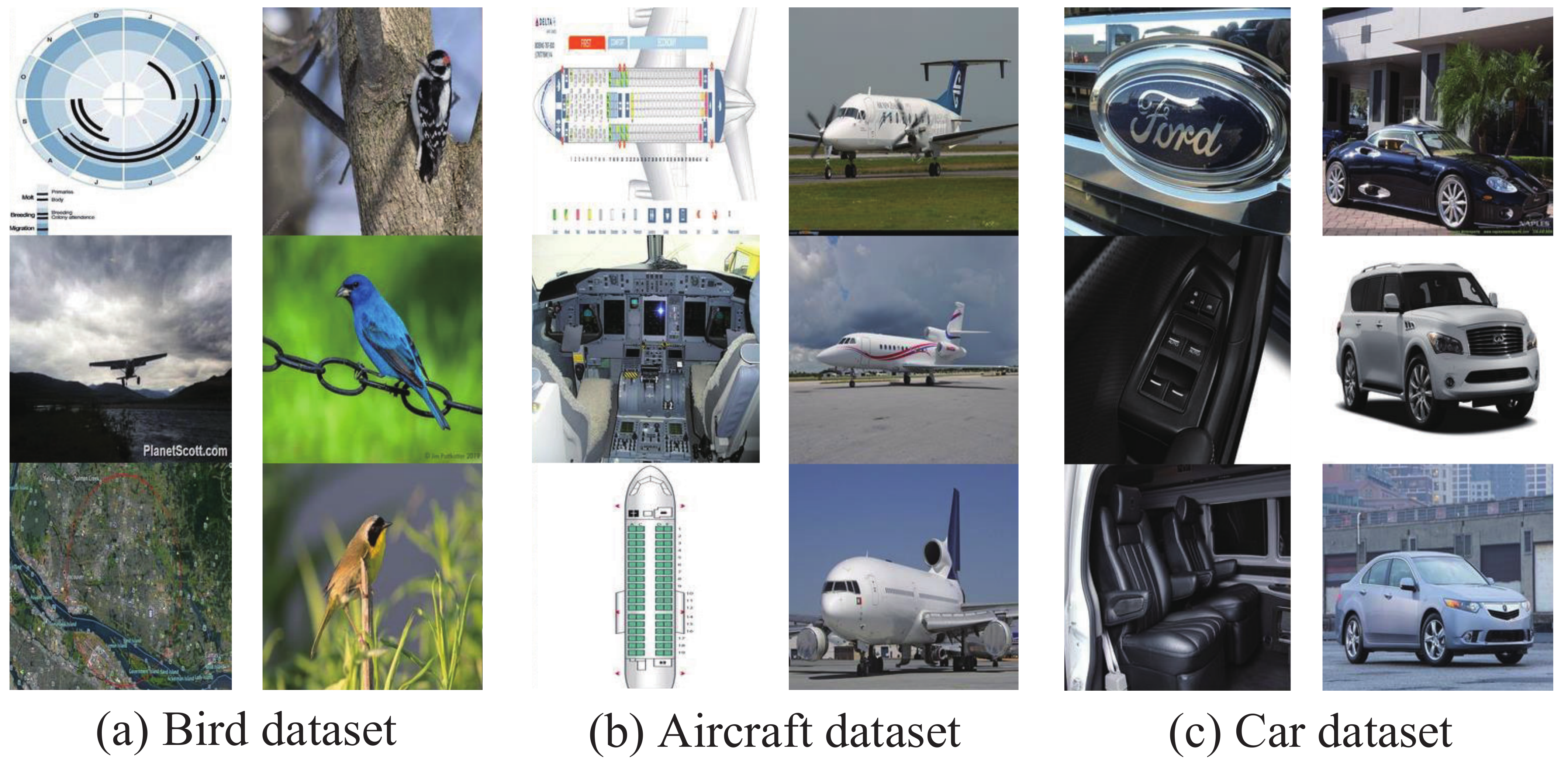} 
	\caption{Sample selection results on three web datasets. For each dataset, the irrelevant noisy and useful images selected by our approach are shown on the left and right, respectively.}
	\label{fig12}
\end{figure}

Fig.~\ref{fig12} visualizes the sample selection results of our approach on three web datasets. From Fig.~\ref{fig12}, we can observe that the irrelevant noisy images and useful images are clearly separated. Most irrelevant noisy images in the bird dataset have no relationship with birds. In the aircraft dataset, irrelevant noisy images are structure charts and cockpits, while in the car dataset, they tend to be logos and internal views of the car. They are related to the aircraft and car but different from images in standard datasets and harmful for training. Although the irrelevant noisy images in these datasets are totally different, our approach is still able to distinguish them from different datasets. These selection results also demonstrate that our approach is robust and can be leveraged to refurbish the web images for practical applications.

{\subsubsection{Summary of Normalization and Label Smoothing}
	
In our experiments, we elaborate on improvements caused by normalization and label smoothing. We first investigate the influence of each module in the performance and output logits. We find that all modules contribute to the final performance, where label smoothing shows a more significant contribution than others. We then demonstrate the advantage of utilizing web images by conducting data augmentation. Next, we illustrate the robustness of our approach through changing backbones. Finally, we analyze the parameter sensitivity and intuitively visualize our sample selection results.}

\section{Conclusion}

In this paper, we presented a simple yet effective training method for web-supervised fine-grained visual classification. Our key idea is to drop irrelevant noisy images and utilize hard examples to boost the model training. Specifically, we select samples that have a large probability cross-entropy as irrelevant noisy images and then drop them during training. Then, we leverage normalization and label smoothing to mine hard examples as well as prevent overfitting. Comprehensive experiments on three real-world scenario datasets demonstrate that our approach is superior to current state-of-the-art methods for web-supervised fine-grained visual classification. {More importantly, our approach is a general and practical method for pure web-supervised learning, which avoids the heavy labor requirements of manual labeling.} It can be applied to other tasks that lack well-labeled training data.


\begin{thebibliography}{10}	
	
	\bibitem{zhang2020deep}
	H. Zhang, Y. Gu, Y. Yao, Z. Zhang, L.~Liu, J. Zhang, and L.	Shao,
	\newblock ``Deep unsupervised self-evolutionary hashing for image retrieval,''
	\newblock {\em IEEE Transactions on Multimedia}, 2020.
	
	\bibitem{hu2020pyretri}
	B. Hu, R. Song, X. Wei, Y. Yao, X. Hua, and Y. Liu,
	\newblock ``Pyretri: A pytorch-based library for unsupervised image retrieval by deep convolutional neural networks,''
	\newblock {\em ACM International Conference on Multimedia}. 4461–--4464, 2020.
	
	\bibitem{wang2020set}
	W. Wang, Y. Shen, H. Zhang, Y. Yao, and L.~Liu,
	\newblock ``Set and rebase: determining the semantic graph connectivity for unsupervised cross modal hashing,''
	\newblock {\em International Joint Conference on Artificial Intelligence}, 853--859, 2020.
	
	\bibitem{yao2019dynamically}
	Y. Yao, Z. Sun, F. Shen, L.~Liu, L. Wang, F. Zhu, L. Ding, G. Wu, and L. Shao,
	\newblock ``Dynamically visual disambiguation of keyword-based image search,''
	\newblock {\em International Joint Conference on Artificial Intelligence}, 996--1002, 2019.
	
	\bibitem{li2020field}
	Z. Li, J. Zhang, Y. Gong, Y. Yao, and Q. Wu,
	\newblock ``Field-wise learning for multi-field categorical data,''
	\newblock {\em International Conference on Neural Information Processing Systems}, 2020.
	
	\bibitem{xie2019attentive}
	G. Xie, L.~Liu, X. Jin, F. Zhu, Z. Zhang, J. Qin, Y. Yao, and L. Shao,
	\newblock ``Attentive region embedding network for zero-shot learning,''
	\newblock {\em IEEE Conference on Computer Vision and Pattern Recognition}, 9384--9393, 2019.
	
	\bibitem{yao2018extracting}
	Y. Yao, J. Zhang, F. Shen, W. Yang, X. Hua, and Z. Tang,
	\newblock ``Extracting privileged information from untagged corpora for classifier learning.,''
	\newblock {\em International Joint Conference on Artificial Intelligence}, 1085--1091, 2018.
	
	\bibitem{xie2020region}
	G. Xie, L.~Liu, F. Zhu, F. Zhao, Z. Zhang, Y. Yao, J. Qin, and L. Shao,
	\newblock ``Region graph embedding network for zero-shot learning,''
	\newblock {\em European Conference on Computer Vision}. 562--580, 2020.
	
	\bibitem{sun2020exploiting}
	Z. Sun, Y. Yao, J. Xiao, L. Zhang, J. Zhang, and Z. Tang,
	\newblock ``Exploiting textual queries for dynamically visual disambiguation,''
	\newblock {\em Pattern Recognition}, 110: 107620, 2021.
	
	\bibitem{zhoutarget}
	T. Zhou, W. Wang, Y. Yao, and J. Shen,
	\newblock ``Target-aware adaptive tracking for unsupervised video object segmentation,''
	\newblock {The DAVIS Challenge on Video Object Segmentation on CVPR Workshop}. 2020.
	
	\bibitem{yao2019tip}
	Y. Yao, F. Shen, J. Zhang, L. Liu, Z. Tang and L. Shao,
	\newblock Extracting Privileged Information for Enhancing Classifier Learning,
	\newblock {IEEE Transactions on Image Processing}, 28(1): 436--450, 2019.
	
	\bibitem{yao2020exploiting}
	Y. Yao, F. Shen, G. Xie, L.~Liu, F. Zhu, J. Zhang, and H. Shen,
	\newblock ``Exploiting web images for multi-output classification: From category to subcategories,''
	\newblock {\em IEEE Transactions on Neural Networks and Learning Systems}, 31(7): 2348--2360, 2020.
	
	\bibitem{zhou2020motion}
	T. Zhou, S. Wang, Y.~Zhou, Y. Yao, J. Li, and L. Shao,
	\newblock ``Motion-attentive transition for zero-shot video object segmentation.,''
	\newblock {\em AAAI Conference on Artificial Intelligence}, 13066--13073, 2020.
	
	\bibitem{chen2020classification}
	T. Chen, J. Zhang, G. Xie, Y. Yao, X. Huang, and Z. Tang,
	\newblock ``Classification constrained discriminator for domain adaptive semantic segmentation,''
	\newblock {\em IEEE International Conference on Multimedia and Expo}, 1--6, 2020.
	
	\bibitem{lu2020hsi}
	J. Lu, H. Liu, Y. Yao, S. Tao, Z. Tang, and J. Lu,
	\newblock ``Hsi road: A hyper spectral image dataset for road segmentation,''
	\newblock {\em IEEE International Conference on Multimedia and Expo}, 1--6, 2020.
	
	\bibitem{luo2019segeqa}
	H. Luo, G. Lin, Z. Liu, F. Liu, Z. Tang, and Y. Yao,
	\newblock ``Segeqa: Video segmentation based visual attention for embodied question answering,''
	\newblock {\em IEEE International Conference on Computer Vision}, 9667--9676, 2019.
	
	\bibitem{imagenet}
	J.~Deng, W.~Dong, R.~Socher, L.-J.~Li, K.~Li, and F.-F.~Li, 
	\newblock ``Imagenet: A large-scale hierarchical image database,''
	\newblock {\em IEEE Conference on Computer Vision and Pattern Recognition}, 248--255, 2009.
	
	\bibitem{zhang2020web}
	C. Zhang, Y. Yao, J. Zhang, J. Chen, P.~Huang, J. Zhang, and Z. Tang,
	\newblock ``Web-supervised network for fine-grained visual classification,''
	\newblock {\em IEEE International Conference on Multimedia and Expo}, 1--6, 2020.
	
	\bibitem{xu2015augmenting}
	Z.~Xu, S.~Huang, Y.~Zhang, and D.~Tao, 
	\newblock ``Augmenting strong supervision using web data for fine-grained categorization,''
	\newblock {\em IEEE International Conference on Computer Vision}, 2524--2532, 2015.	
	
	\bibitem{niu2018webly}
	L.~Niu, A.~Veeraraghavan, and A.~Sabharwal, 
	\newblock ``Webly supervised learning meets zero-shot learning: A hybrid approach for fine-grained classification,''
	\newblock {\em IEEE Conference on Computer Vision and Pattern Recognition}, 7171--7180, 2018.
	
	\bibitem{cui2016fine}
	Y.~Cui, F.~Zhou, Y.~Lin, and S.~Belongie, 
	\newblock ``Fine-grained categorization and dataset bootstrapping using deep metric learning with humans in the loop,''
	\newblock {\em IEEE Conference on Computer Vision and Pattern Recognition}, 1153--1162, 2016.
	
	\bibitem{yao2019towards}
	Y. Yao, J. Zhang, F. Shen, L.~Liu, F. Zhu, D. Zhang, and H. Shen,
	\newblock ``Towards automatic construction of diverse, high-quality image datasets,''
	\newblock {\em IEEE Transactions on Knowledge and Data Engineering}, 32(6): 1199--1211, 2020.	
	
	\bibitem{yao2016automatic}
	Y. Yao, J. Zhang, F. Shen, X. Hua, J. Xu, and Z. Tang,
	\newblock ``Automatic image dataset construction with multiple textual metadata,''
    \newblock {\em IEEE International Conference on Multimedia and Expo}, 1--6, 2016.
    
    \bibitem{schroff2011harvesting}
    F.~Schroff, A.~Criminisi, and A.~Zisserman,
    \newblock ``Harvesting image databases from the web,''
    \newblock {\em IEEE Transactions on Pattern Analysis and Machine Intelligence}, 33(4): 754--766, 2011.
    
    \bibitem{arpit2017closer}
    D.~Arpit, S.~Jastrz{\k{e}}bski, N.~Ballas, D.~Krueger, E.~Bengio, M.~S. Kanwal, T.~Maharaj, A.~Fischer, A.~Courville, Y.~Bengio, and S.~Lacoste-Julien,
    \newblock ``A closer look at memorization in deep networks,''
    \newblock {\em International Conference on Machine Learning}, 233--242, 2017.
    
    \bibitem{zhang2016understanding}
    C.~Zhang, S.~Bengio, M.~Hardt, B.~Recht, and O.~Vinyals,
    \newblock ``Understanding deep learning requires rethinking generalization,''
    \newblock {\em International Conference on Learning Representations}, 1--15, 2016.
    
    \bibitem{reed2014training}
    S.~Reed, H.~Lee, D.~Anguelov, C.~Szegedy, D.~Erhan, and A.~Rabinovich,
    \newblock ``Training deep neural networks on noisy labels with bootstrapping,''
    \newblock {\em arXiv}, 1412.6596, 2014.	
	
	\bibitem{yao2018discovering}
	Y. Yao, J. Zhang, F. Shen, W. Yang, P.~Huang, and Z. Tang,
	\newblock ``Discovering and distinguishing multiple visual senses for polysemous words,''
	\newblock {\em AAAI Conference on Artificial Intelligence}, 523--530, 2018.
	
	\bibitem{goldberger2016training}
	J.~Goldberger and E.~Ben-Reuven,
	\newblock ``Training deep neural-networks using a noise adaptation layer,''
	\newblock {\em International Conference on Learning Representations}, 1--9, 2016.
	
	\bibitem{malach2017decoupling}
	E.~Malach and S.~Shalev-Shwartz,
	\newblock ``Decoupling "when to update" from "how to update",''
	\newblock {\em The Conference and Workshop on Neural Information Processing Systems}, 960--970, 2017.		
	
	\bibitem{yao2019tmm}
	Y.~Yao, F.~Shen, J.~Zhang, L. Liu, Z. Tang and L. Shao,
	\newblock ``Extracting Multiple Visual Senses for Web Learning,''
	\newblock {\em IEEE Transactions on Multimedia}, 21(1): 184--196, 2019.
	
	\bibitem{yao2017exploiting}
	Y.~Yao, J.~Zhang, F.~Shen, X.~Hua, J.~Xu, and Z.~Tang,
	\newblock ``Exploiting web images for dataset construction: A domain robust approach,''
	\newblock {\em IEEE Transactions on Multimedia}, 19(8): 1771--1784, 2017.
	
	\bibitem{ding2020approximate}
	L. Ding, S. Liao, Y. Liu, L.~Liu, F. Zhu, Y. Yao, L. Shao, and X. Gao,
	\newblock ``Approximate kernel selection via matrix approximation,''
	\newblock {\em IEEE Transactions on Neural Networks and Learning Systems}, 31(11): 4881--4891, 2020.
	
	\bibitem{yao2016domain}
	Y. Yao, X. Hua, F. Shen, J. Zhang, and Z. Tang,
	\newblock ``A domain robust approach for image dataset construction,''
	\newblock {\em ACM international conference on Multimedia}, 212--216, 2016.		
	
	\bibitem{han2018co}
	B.~Han, Q.~Yao, X.~Yu, G.~Niu, M.~Xu, W.~Hu, I.~Tsang, and M.~Sugiyama,
	\newblock ``Co-teaching: Robust training of deep neural networks with extremely noisy labels,''
	\newblock {\em The Conference on Neural Information Processing Systems}, 8527--8537, 2018.
	
	\bibitem{wang2018iterative}
	Y.~Wang, W.~Liu, X.~Ma, J.~Bailey, H.~Zha, L.~Song, and S.-T.~Xia,
	\newblock ``Iterative learning with open-set noisy labels,''
	\newblock {\em IEEE Conference on Computer Vision and Pattern Recognition}, 8688--8696, 2018.
	
	\bibitem{liang2017enhancing}
	S.~Liang, Y.~Li, and R.~Srikant,
	\newblock ``Enhancing the reliability of out-of-distribution image detection in neural networks,''
	\newblock {\em arXiv}, 1706.02690v4, 2017.
	
	\bibitem{l2softmax}
	R.~Ranjan, C.~D. Castillo, and R.~Chellappa, 
	\newblock ``L2-constrained softmax loss for discriminative face verification,''
	\newblock {\em arXiv}, 1703.09507, 2017.
	
	\bibitem{amsoftmax}
	F.~Wang, J.~Cheng, W.~Liu, and H.~Liu, 
	\newblock ``Additive margin softmax for face verification,'' 
	\newblock {\em IEEE Signal Processing Letters}, 25(7): 926--930, 2018.
	
	\bibitem{hinton2015distilling}
	G.~Hinton, O.~Vinyals, and J.~Dean, 
	\newblock ``Distilling the Knowledge in a Neural Network,'' 
	\newblock {\em arXiv}, 1503.02531v1, 2015.
	
	\bibitem{szegedy2016rethinking}
	C.~Szegedy, V.~Vanhoucke, S.~Ioffe, J.~Shlens, and Z.~Wojna,
	\newblock ``Rethinking the inception architecture for computer vision,''
	\newblock {\em IEEE Conference on Computer Vision and Pattern Recognition}, 2818--2826, 2016.
	
	\bibitem{zhuang2017attend}
	B.~Zhuang, L.-Q.~Liu, Y.~Li, C.-H.~Shen, and I.~Reid,
	\newblock ``Attend in groups: a weakly-supervised deep learning framework for learning from web data,''
	\newblock {\em IEEE Conference on Computer Vision and Pattern Recognition}, 1878--1887, 2017.
	
	\bibitem{AAAI2020}
	C.~Zhang, Y.~Yao, H.~Liu, G.~S. Xie, X.~Shu, T.~Zhou, Z.~Zhang, F.~Shen, and Z.~Tang,
	\newblock ``Web-supervised network with softly update-drop training for fine-grained visual classification,''
	\newblock {\em AAAI Conference on Artificial Intelligence}, 12781--12788, 2020.
	
	\bibitem{huang2016part}
	S.~Huang, Z.~Xu, D.~Tao, and Y.~Zhang,
	\newblock ``Part-stacked cnn for fine-grained visual categorization,''
	\newblock {\em IEEE Conference on Computer Vision and Pattern Recognition}, 1173--1182, 2016.
	
	\bibitem{yao2016coarse}
	H.~Yao, S.~Zhang, Y.~Zhang, J.~Li, and Q.~Tian,
	\newblock ``Coarse-to-fine description for fine-grained visual categorization,''
	\newblock {\em IEEE Transactions on Image Processing}, 25(10): 4858--4872, 2016.
	
	\bibitem{lam2017fine}
	M.~Lam, B.~Mahasseni, and S.~Todorovic,
	\newblock ``Fine-grained recognition as hsnet search for informative image parts,''
	\newblock {\em IEEE Conference on Computer Vision and Pattern Recognition}, 2520--2529, 2017.
	
	\bibitem{wei2018mask}
	X.~S. Wei, C.~W. Xie, J.~Wu, and C.~Shen,
	\newblock ``Mask-CNN: Localizing parts and selecting descriptors for fine-grained bird species categorization,''
	\newblock {\em Pattern Recognition}, 76: 704--714, 2018.
	
	\bibitem{zhang2014part}
	N.~Zhang, J.~Donahue, R.~Girshick, and T.~Darrell,
	\newblock ``Part-based R-CNNs for fine-grained category detection,''
	\newblock {\em European Conference on Computer Vision}, 834--849, 2014.	
	
	\bibitem{yao2020bridging}
	Y.~Yao, X.~Hua, G. Gao, Z. Sun, Z. Li, and J. Zhang,
	\newblock ``Bridging the web data and fine-grained visual recognition via alleviating label noise and domain mismatch,''
	\newblock {\em ACM International Conference on Multimedia}, 1735--1744, 2020.
	
	\bibitem{sun2020crssc}
	Z. Sun, X. Hua, Y. Yao, X. Wei, G. Hu, and J. Zhang,
	\newblock ``Crssc: salvage reusable samples from noisy data for robust learning,''
	\newblock {\em ACM International Conference on Multimedia}, 92--101, 2020.
	
	\bibitem{zhang2020data}
	C. Zhang, Y. Yao, X. Shu, Z. Li, Z. Tang, and Q. Wu,
	\newblock ``Data-driven meta-set based fine-grained visual recognition,''
	\newblock {\em ACM International Conference on Multimedia}, 2372--2381, 2020.
	
	\bibitem{mnet}
	L.~Jiang, Z.-Y.~Zhou, T.~Leung, L.-J.~Li, and F.-F.~Li,
	\newblock ``Mentornet: Learning data-driven curriculum for very deep neural networks on corrupted labels,''
	\newblock {\em International Conference on Machine Learning}, 1--20, 2017.
	
	\bibitem{pencil2019}
	K.~Yi and J,-X.~Wu,
	\newblock ``Probabilistic End-to-end Noise Correction for Learning with Noisy Labels,''
	\newblock {\em IEEE Conference on Computer Vision and Pattern Recognition}, 7017--7025, 2019.
		
	\bibitem{patrini}
	G.~Patrini, A.~Rozza, A.~Krishna-Menon, R.~Nock, and L.-Z.~Qu,
	\newblock ``Making deep neural networks robust to label noise: A loss correction approach,''
	\newblock {\em IEEE Conference on Computer Vision and Pattern Recognition}, 1944--1952, 2017.
	
	\bibitem{fu2017look}
	J.~Fu, H.~Zheng, and T.~Mei,
	\newblock ``Look closer to see better: Recurrent attention convolutional neural network for fine-grained image recognition,''
	\newblock {\em IEEE Conference on Computer Vision and Pattern Recognition}, 4438--4446, 2017.
	
	\bibitem{lin2015bilinear}
	T.-Y.~Lin, A.~RoyChowdhury, and S.~Maji,
	\newblock ``Bilinear cnn models for fine-grained visual recognition,''
	\newblock {\em IEEE International Conference on Computer Vision}, 1449--1457, 2015.
	
	\bibitem{zheng2017learning}
	H.~Zheng, J.~Fu, T.~Mei, and J.~Luo,
	\newblock ``Learning multi-attention convolutional neural network for fine-grained image recognition,''
	\newblock {\em IEEE International Conference on Computer Vision}, 5209--5217, 2017.
	
	\bibitem{wang2018learning}
	Y.~Wang, V.-I.~Morariu, and L.-S.~Davis,
	\newblock ``Learning a discriminative filter bank within a cnn for fine-grained recognition,''
	\newblock {\em IEEE Conference on Computer Vision and Pattern Recognition}, 4148--4157, 2018.
	
	\bibitem{ge2019weakly}
	W.~Ge, X.~Lin, and Y.~Yu,
	\newblock ``Weakly supervised complementary parts models for fine-grained image classification from the bottom up,''
	\newblock {\em IEEE Conference on Computer Vision and Pattern Recognition}, 3034--3043, 2019.
	
	\bibitem{zheng2019looking}
	H.~Zheng, J.~Fu, Z.-J. Zha, and J.~Luo,
	\newblock ``Looking for the devil in the details: Learning trilinear attention sampling network for fine-grained image recognition,''
	\newblock {\em IEEE Conference on Computer Vision and Pattern Recognition}, 5012--5021, 2019.
	
	\bibitem{chen2019destruction}
	Y.~Chen, Y.~Bai, W.~Zhang, and T.~Mei,
	\newblock ``Destruction and construction learning for fine-grained image recognition,''
	\newblock {\em IEEE Conference on Computer Vision and Pattern Recognition}, 5157--5166, 2019.
	
	\bibitem{korsch2019classification}
	D.~Korsch, P.~Bodesheim, and J.~Denzler,
	\newblock ``Classification-specific parts for improving fine-grained visual categorization,''
	\newblock {\em arXiv}, 1909.07075, 2019.

	\bibitem{xu2016webly}
	Z.~Xu, S.~Huang, Y.~Zhang, and D.~Tao,
	\newblock ``Webly-supervised fine-grained visual categorization via deep domain adaptation,''
	\newblock {\em IEEE Transactions on Pattern Analysis and Machine Intelligence}, 40(5): 1100--1113, 2016.
	
	\bibitem{niu2015visual}
	L.~Niu, W.~Li, and D.~Xu,
	\newblock ``Visual recognition by learning from web data: A weakly supervised domain generalization approach,''
	\newblock {\em IEEE Conference on Computer Vision and Pattern Recognition}, 2774--2783, 2015.
	
	\bibitem{xiao2015learning}
	T.~Xiao, T.~Xia, Y.~Yang, C.~Huang, and X.~Wang,
	\newblock ``Learning from massive noisy labeled data for image classification,''
	\newblock {\em IEEE Conference on Computer Vision and Pattern Recognition}, 2691--2699, 2015.
	
	\bibitem{cosineloss}
	Y.~Liu, H.~Li, and X.~Wang,
	\newblock ``Learning deep features via congenerous cosine loss for person recognition,''
	\newblock {\em arXiv}, 1702.06890, 2017.
	
	\bibitem{Normface}
	F.~Wang, X.~Xiang, J.~Cheng, and A.-L.~Yuille,
	\newblock ``Normface: $l_2$ hypersphere embedding for face verification,''
	\newblock {\em ACM International Conference on Multimedia}, 1041--1049, 2017.

	\bibitem{cui2018large}
	Y.~Cui, Y.~Song, C.~Sun, A.~Howard, and S.~Belongie,
	\newblock ``Large scale fine-grained categorization and domain-specific transfer learning,''
	\newblock {\em IEEE Conference on Computer Vision and Pattern Recognition}, 4109--4118, 2018.
	
	\bibitem{wah2011caltech}
	C.~Wah, S.~Branson, P.~Welinder, P.~Perona, and S.~Belongie,
	\newblock ``The caltech-ucsd birds-200-2011 dataset,''
	\newblock {\em Technical Report}, CNS-TR-2011-001, California Institute of Technology, 2011.
	
	\bibitem{aircraft}
	S.~Maji, E.~Rahtu, J.~Kannala, M.~Blaschko, and A.~Vedaldi,
	\newblock ``Fine-grained visual classification of aircraft,''
	\newblock {\em arXiv}, 1306.5151, 2013.
	
	\bibitem{car196}
	J.~Krause, M.~Stark, J.~Deng, and L.~Fei-Fei,
	\newblock ``3D object representations for fine-grained categorization,''
	\newblock {\em IEEE International Conference on Computer Vision}, 554--561, 2013.
	
\end{thebibliography}
\end{document}